\documentclass[pdflatex,sn-mathphys-num]{sn-jnl}%
 \usepackage{url,hyperref} 
\usepackage[utf8]{inputenc}
\usepackage{pifont}
\usepackage{enumitem}
\usepackage{array,tikz}
\usepackage{graphicx}%
\usepackage{multirow}%
\usepackage{amsmath,amssymb,amsfonts}%
\usepackage{amsthm}%
\usepackage{mathrsfs}%
\usepackage[title]{appendix}%
\usepackage{xcolor}%
\usepackage{textcomp}%
\usepackage{manyfoot}%
\usepackage{booktabs}%
\usepackage{algorithm}%
\usepackage{algorithmicx}%
\usepackage{algpseudocode}%
\usepackage{listings}%
\begin{document}

\title[Article Title]{Generative to Agentic AI: Survey, Conceptualization, and Challenges }


\author*[1]{\fnm{Johannes} \sur{Schneider}}\email{johannes.schneider@uni.li}
\affil[1]{University of Liechtenstein, Vaduz, Liechtenstein}

\abstract{
Agentic AI builds upon Generative AI (GenAI). It constitutes the next major step in the evolution of AI with much stronger reasoning, learning, and interaction capabilities that enable more autonomous behavior to tackle complex, long-running tasks. Since the initial release of ChatGPT 3.5, Generative AI has seen widespread adoption, giving users firsthand experience. However, the distinction between Agentic AI and GenAI remains less well understood. To address this gap, our survey is structured in two parts. In the first part, we compare GenAI and Agentic AI using existing literature, discussing their key characteristics, how Agentic AI remedies limitations of GenAI, and the major steps in GenAI's evolution toward Agentic AI. We also distinguish between early Agentic AI aiming at reasoning and interacting and the increasingly emerging paradigm of self-improving Agentic AI that operates continuously and learns from its experiences by updating itself. This section is intended for a broad audience, including academics in both social sciences and engineering, as well as industry professionals. It provides the necessary insights to comprehend novel applications that are possible with Agentic AI but not with GenAI. In the second part, we deep dive into novel aspects of Agentic AI, including recent developments and practical concerns such as defining agents. Finally, we discuss several challenges that could serve as a future research agenda, while cautioning against risks that can emerge when exceeding human intelligence.}



\maketitle


\section{Introduction} \label{secintro}
\begin{figure*}[h]
\centering
\includegraphics[width=\textwidth]{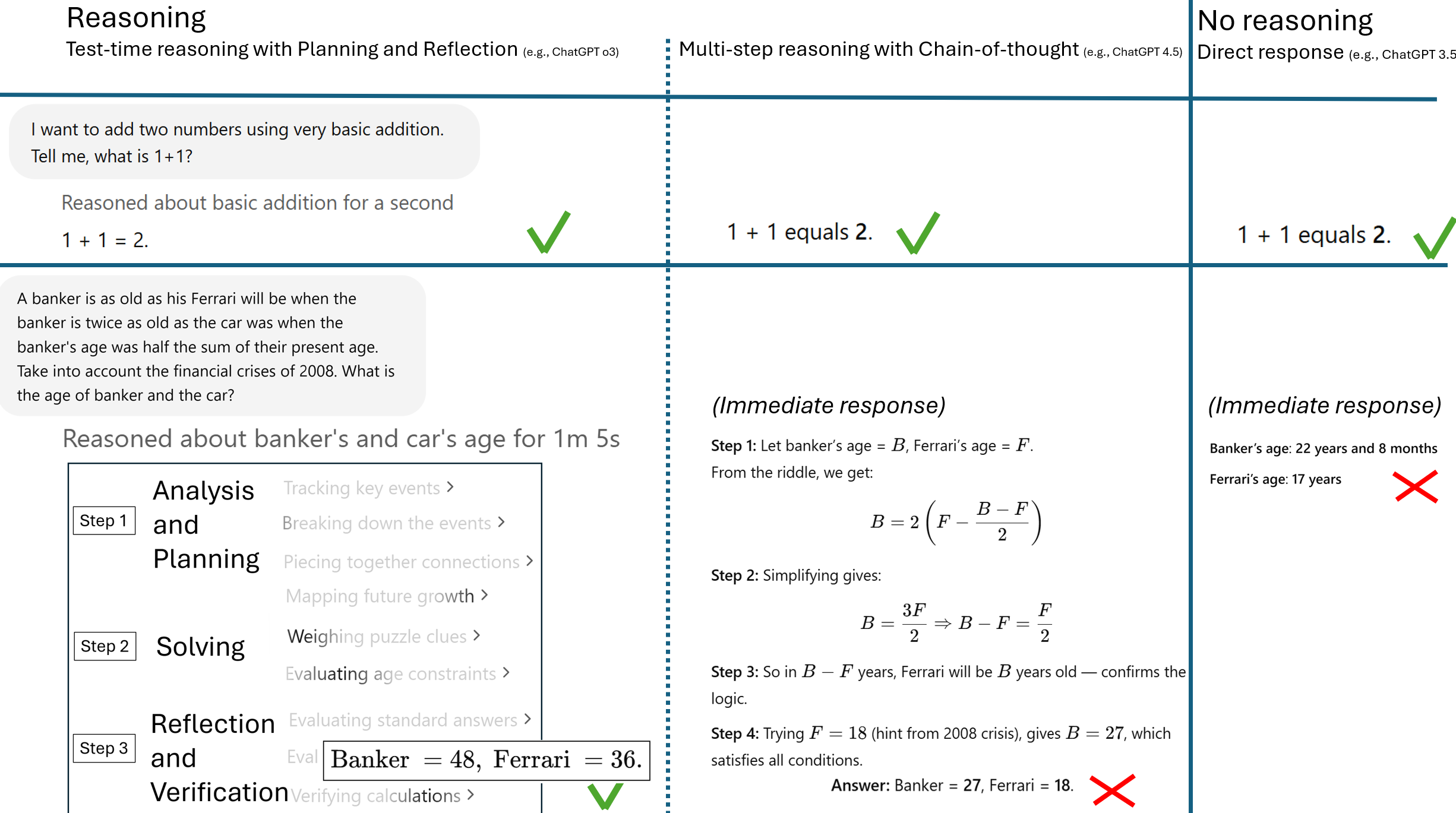}
\caption{Reasoning models perform extensive problem-dependent computations, commonly employing problem analysis, planning and reflection, while non-reasoning is shown by immediate responses without intermediate steps.}\label{fig:reasEx}
\end{figure*}

Agentic AI constitutes a \textbf{paradigm shift in artificial intelligence}, enabling systems to act independently, pursue broad objectives rather than isolated decisions, and carry out complex tasks that require reasoning elements such as planning and reflection. Agents can act proactively rather than just reactively. While Agentic AI builds on earlier seeds of shallow reasoning and interaction already observed in early GenAI systems, it extends and structures these capabilities profoundly. Agents (i) interact with environments and tools, and (ii) perform deep reasoning, which manifests as multi-step, problem-dependent computation with planning and reflection, and increasingly (iii) agents learn from collected experiences to better understand their environment and improve on (recurring) tasks. For example, reasoning models such as ChatGPT o3 or ChatGPT 6 Astra in ``Pro'' mode can spend minutes to hours processing user and self-generated prompts as part of a search and planning process, whereas earlier GenAI models like ChatGPT 3.5 (released in 2022) typically provided instantaneous responses, as shown in Figure \ref{fig:reasEx}. Increasingly, harnesses steering AI models trigger processes to store, distill, and retrieve experiences, allowing agents to learn during long-running tasks as observed for coding and personal agent (harnesses) operating 24/7 \cite{liu2026dive, opcl26sel}.

Beyond limited reasoning capabilities, tasks handled by GenAI are typically structured to be solved directly using the available information—that is, by generating an output from the input without engaging with the environment or external tools. In contrast, Agentic AI systems integrate elements of reinforcement learning: Agents interact with environments using tools through a sequence of actions, making observations and receiving feedback that informs and guides future actions via learning.
\begin{figure*}[h]
\centering
\includegraphics[width=0.8\textwidth]{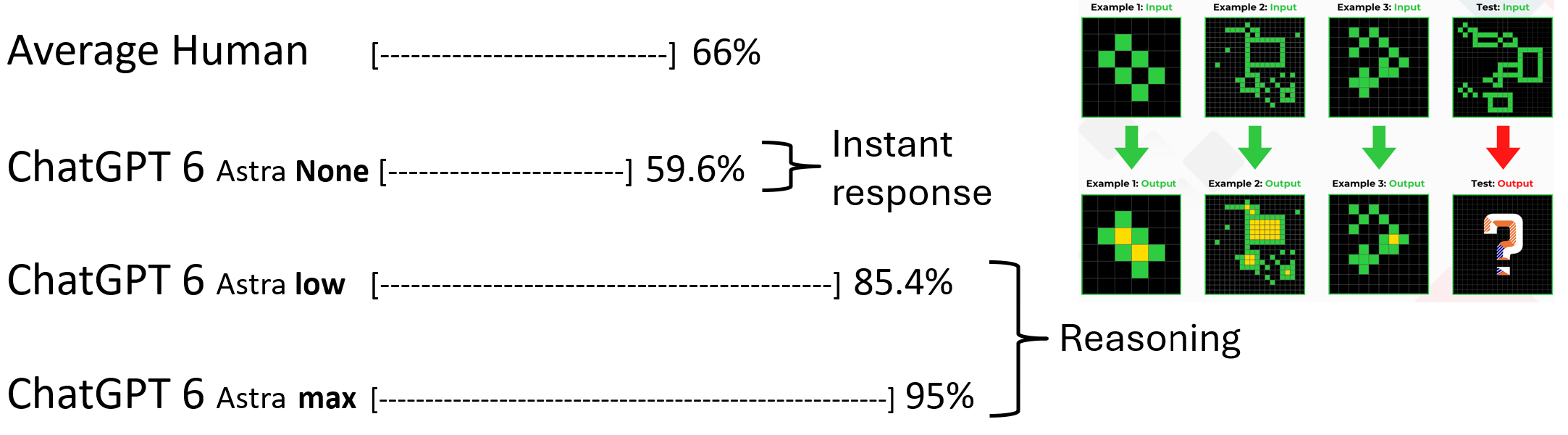}
\caption{On the ARC challenge~\citep{cho19arc}, Agentic AI with growing reasoning effort involving planning and reflection dramatically outperform.}\label{fig:arc}
\end{figure*} 
This new wave of AI fosters \textbf{novel opportunities and challenges}, making it crucial to understand and conceptualize differences clearly. Defining and using an autonomous agent is more challenging and riskier, as agents have a greater direct real-world impact: they can directly alter the environment by taking actions and solve tasks with less-detailed instructions, leading to more diverse, unpredictable, and harder-to-control solutions. Their dangers have been demonstrated by multiple unintendet hacker attacks, e.g., on HuggingFace by OpenAI's flagship model \cite{op26hug}. Moreover, capabilities such as memory use, tool selection, and open-ended exploration amplify both the potential benefits and risks compared to GenAI. On the positive side, AI agents, like GenAI, can be used with plain, everyday language instead of complex programming languages, boosting ease of adoption. On the negative side, properly specifying an agent to account for various eventualities is more demanding than crafting a prompt for a narrower task.
While some experts already warned against the threats of GenAI~\citep{hea23}, many researchers deemed those threats as overstated~\citep{bus24}. This is not a surprise, as GenAI models struggled to solve tasks that were easy for laypeople. For instance, in the ARC challenge~\citep{cla18}—which consists of tasks considered simple for ordinary people—early GenAI models such as GPT 4o  achieved below 10\%. However, reasoning models that are part of \textbf{Agentic AI show striking improvements}, as illustrated in Figure \ref{fig:arc}. In other areas, such as programming~\citep{el25com} and generating coherent and accurate radiology reports, Agentic AI with reasoning capabilities also outperform other models~\citep{zho24ev}. These outcomes also suggest that Agentic AI represents a major step toward artificial general intelligence (AGI), where AI is not confined to a narrow domain but can generalize to new situations. However, scaling compute at test-time alone (a key aspect of ``deep'' reasoning) may not guarantee progress toward AGI, and important challenges remain, including training data limitations and error accumulation.
Thus, the debate on the risks and opportunities of AI should intensify again, requiring a renewed and cautious assessment. Understanding the differences between the two is important for businesses to choose the right technology to save time and reduce costs, and for society and policymakers to engage in a meaningful ethical and social debate surrounding the technology~\citep{mar25gen}.

However, the situation becomes more intricate as \textbf{Agentic AI is not a magic bullet} for all problems. In particular, it does not offer dramatic improvements over GenAI in all areas. This makes it challenging to assess which problems benefit from clear-cut gains that enable novel applications. Like GenAI, Agentic AI can make mistakes on relatively simple problems~\citep{zho24ev}. Social intelligence is not necessarily higher in models performing deep reasoning~\citep{hou24en}, although differences may emerge in tasks requiring the anticipation of behavioral strategies, such as those found in games. Only minor improvements using Agentic AI have been observed on translation tasks~\citep{che25ev}. On the well-known MMLU benchmark—consisting of multiple-choice questions spanning factual and procedural knowledge from various domains—reasoning models show only a slight edge, as illustrated in Figure \ref{fig:mmlu}, and at the cost of significantly more computation. Many of these questions are considered ``hard'' in the sense that they require PhD level knowledge in specific domains. Likely, there are a number of analogous questions to those in the benchmark in the model's training data -- in the most extreme case even the questions and answers themselves. Thus, the questions may be easier to answer through association rather than genuine reasoning. Generally, fine-tuning GenAI models for specific domains may outperform Agentic AI counterparts that lack domain-specific knowledge, while also incurring lower computational costs.
Up to 2025, the MMLU benchmark was used extensively to compare large language model (LLM) performance. Now, it is mostly solved and its harder version (MMLU-Pro) is saturated, i.e., most frontier models score around 90\% with relatively small gaps. 

\begin{figure}
    \centering
    \includegraphics[width=0.9\linewidth]{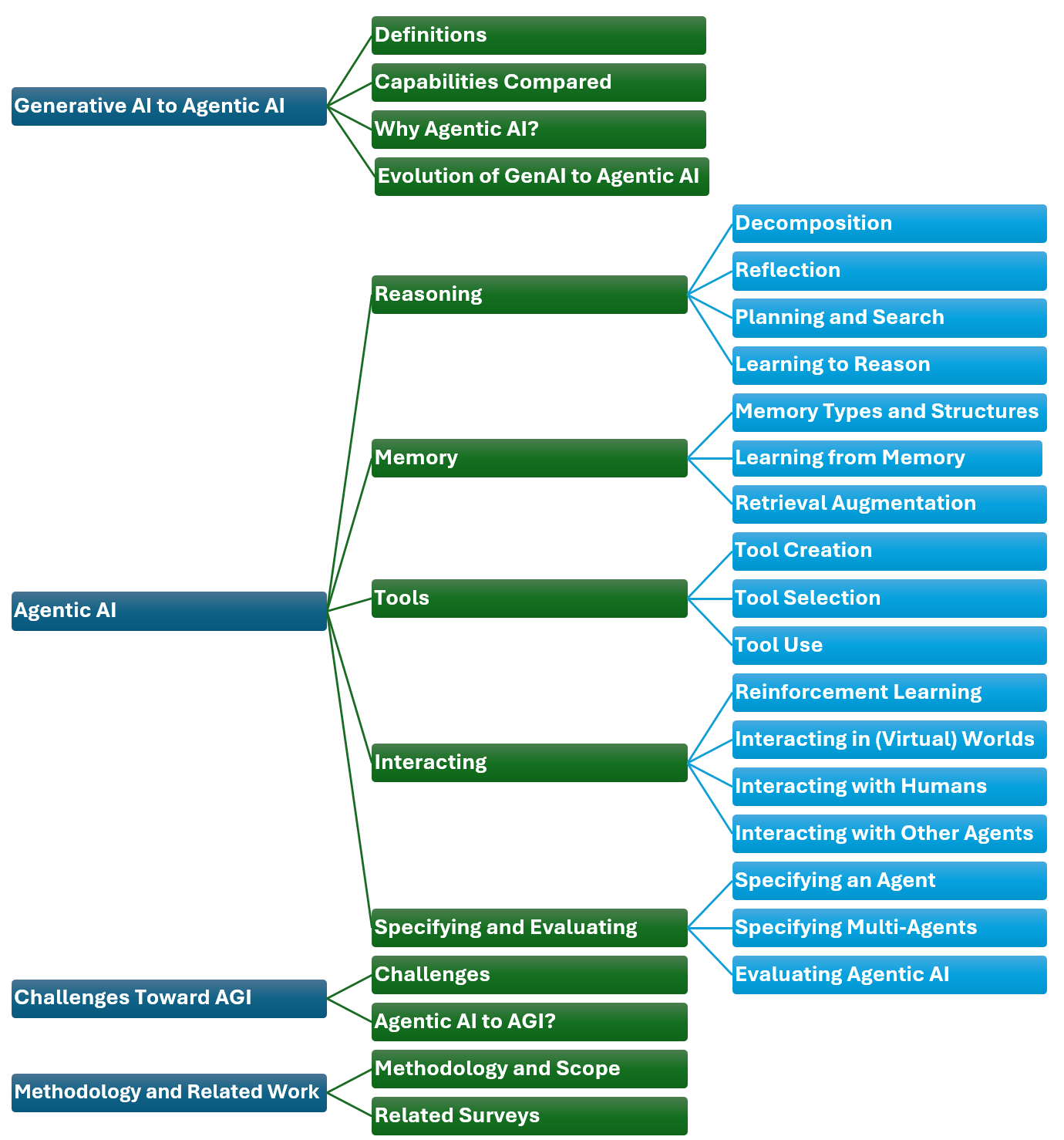}
    \caption{Overview: The high-level discussion of GenAI to Agentic AI is followed by an in-depth treatment of Agentic AI, followed by challenges and outlook}
    \label{fig:outline}
\end{figure}

\begin{figure*}[h]
\centering
\includegraphics[width=\textwidth]{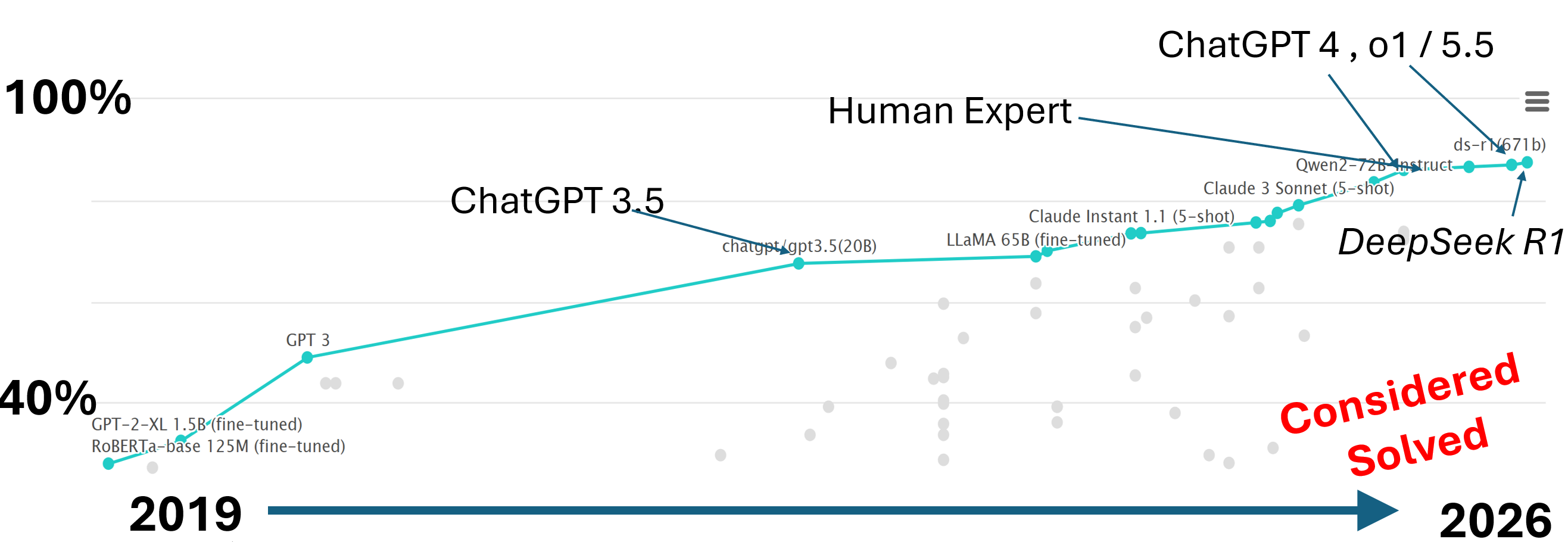}
\caption{Agentic AI and GenAI performs similarly on MMLU~\citep{hen21mml} measuring a wide range of capabilities across disciplines(data from~\citep{pap25sot}). The benchmark is considered ``solved''.}\label{fig:mmlu}
\end{figure*}

\textbf{Contribution and Overview:} This manuscript enables the reader to better understand the novel risks and opportunities of Agentic AI through various conceptualizations and contextualizations. As detailed in Figure \ref{fig:outline}, we first provide definitions contrasting GenAI and Agentic AI and compare their high-level capabilites. We also elaborate on limitations of GenAI and how Agentic AI overcomes (some of) them. While Agentic AI is often portrayed as an unforeseeable, sudden change causing instant disruption, we document its evolution from early GenAI systems, highlighting key milestones related to reasoning and interaction, and noting overlaps and differences with well-established AI paradigms such as reinforcement learning. Next, we provide a deep dive into Agentic AI. We discuss key aspects of Agentic AI, including reasoning (decomposition, multi-step reasoning, reflection, search and planning, memory, tools and interaction with the environment covering also a range of applications). We elaborate on how to specify AI agents, including both single-agent and multi-agent systems as well as how to evaluate Agentic AI systems. Finally, we describe the challenges of Agentic AI to further progress towards AGI, which constitute research opportunities. We also caution on potential risk coming with AGI. Before concluding we elaborate on our research methodology and related surveys.




\begin{figure*}[h]
\centering
\includegraphics[width=\textwidth]{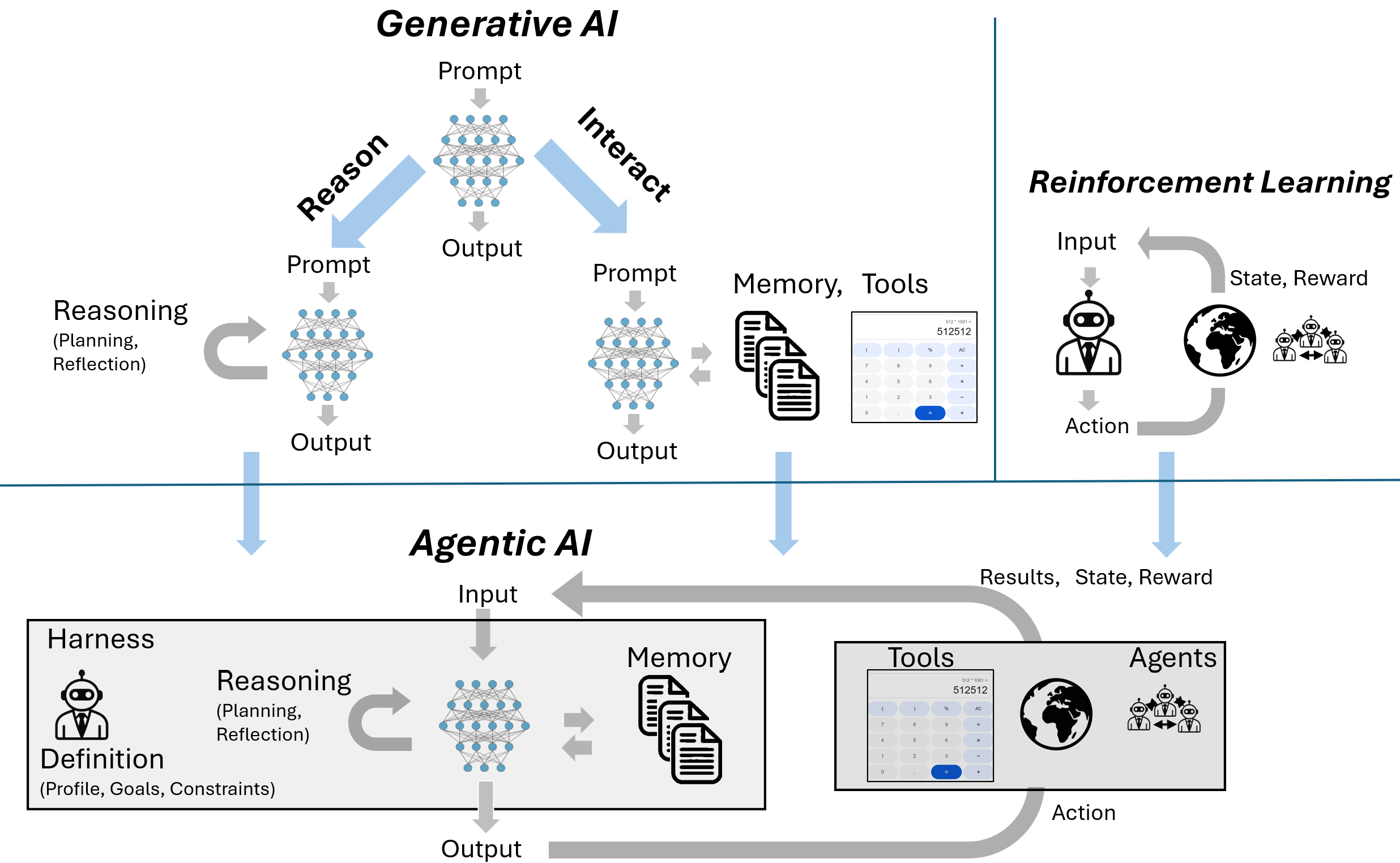}
\caption{From Generative AI to Agentic AI. Early GenAI such as GPT-3~\citep{bro20} showed with minor effort basic reasoning and tool usage capabilities. These became much more profound over time, in particular, with Agentic AI, which includes elements from reinforcement learning such as interacting with the environment and other agents. Also the harness coordinator between memory, tools and the environment became more elaborate.}\label{fig:gentoAge}
\end{figure*}


\section{Generative AI to Agentic AI}
We start with defining both GenAI and Agentic AI before elaborating on their key characteristics, discussing the need for Agentic AI and the evolution of GenAI to Agentic AI from a research perspective.

\subsection{Defining GenAI and Agentic AI}
While many academic works have some notion of Agentic AI, we found that leading AI companies provided most concise definitions covering different angles:

\noindent \emph{``Agentic AI uses sophisticated reasoning and iterative planning to autonomously solve complex, multi-step problems''} (NVidia~\citep{pou24ag}).

\noindent \emph{``Agentic AI systems are AI systems that can pursue complex goals with limited direct supervision''} (OpenAI~\citep{sha23ag}).

NVidia's definition focuses on what AI does and its capabilities (reasoning and planing), emphasizing independence in execution to solve given problems. OpenAI's definition does not restrict the mode of operation and is broader. It emphasizes independence in direction to reach goals, i.e., the agent may choose its own problems to solve.

We emphasize differences to GenAI. We start with a short definition focusing on high-level, non-technical aspects:
\noindent \emph{``Generative AI generates and transforms content based on specific user instructions.''}\\
\noindent \emph{``Agentic AI can autonomously pursue goals in dynamic environments requiring adaptation, interaction and reasoning.''}
Our definition stresses independence to pursue goals (as OpenAI), mentions the how (as NVidia), and emphasizes the requirements that arise in environments these agents are expected to operate in. Our definition emphasizes differences in characteristics as shown in Table \ref{tab:compCap}.

A more comprehensive definition with a more academic phrasing:\\
\noindent \emph{Definition of Generative AI: ``A system based on a foundation model that generates digital artifacts based on natural user instructions.''}\\
Digital artifacts may range from simple binary decisions to text, images, and videos. Natural user instructions refers to instructions that are easy to understand for humans and do not require deep technical or specific knowledge of the AI system, e.g., human language, or visual depiction such as photos or sketches.\\
\noindent \emph{Definition of Agentic AI: ``A system based on a foundation model that performs tasks potentially yielding artifacts based on natural user instructions, where the system is able to conduct and express complex reasoning including planning and reflection to pursue goals and solve tasks that require interaction with an environment and elaborate tool use.''}\\
Agentic AI can go beyond creating digital artifacts, as it might, e.g., control robots to produce or alter physical objects. It might also execute tasks that do not yield any directly observable artifact for a user, e.g., smart grid management or computer internal system optimizations such as scheduling computational resources to minimize costs while obeying constraints. That is, while an agent takes action that impact our world, the outcome is not a tangible artifact such as a text or images. In both definitions we included the term foundation model, which is a large deep learning network~\citep{schn24f} though agents could be (and have been) built with other model types. However,  many general and more detailed, technical aspects expressed in our and other works assume the usage of a foundation model as otherwise many claims are not substantiated.

When emphasizing the interconnection between prior AI developments, one might say that:\\
\noindent \emph{``Agentic AI builds on Generative AI by combining foundation models enhanced with the capability for tool usage, memory access with reinforcement learning with its notion of agents and planning to enable interaction and reasoning.''}
In contrast, GenAI only relies on foundation models with early models having only limited capabilities for tool usage and reasoning - This definition is aligned with Figure \ref{fig:gentoAge}.

A foundation model is to GenAI what an agent is to Agentic AI. Agents themselves are systems including a foundation model (see Figure \ref{fig:gentoAge}). However, this is somewhat oversimplified, as the construct of agents in AI is rather old and goes beyond agents being a technical component~\citep{rus21}. Agentic AI is a subfield of AI, while an AI agent is the central object of study. Thus, Agentic AI is more comprehensive including procedures for training, evaluating, defining agents, and coordinating multiple agents. It also covers non-technical aspects such as ethical, economic, social and philosophical debates. For brevity we shall often omit the term ``systems'' in combination with GenAI and Agentic AI.

Also we characterize the emerging paradigm of self-improving Agentic AI briefly as:
\noindent \emph{``Self-improving Agentic AI adds continual learning to Agentic AI.''}
Self-improving agents possess the ability to alter their knowledge base persistently. This goes beyond simply maintaining past interactions as part of the prompt. Early works demonstrated that examples of tasks (with solution processes) improved performance, e.g., few-shot prompting enabling in-context learning \cite{bro20}. Thus, past experiences \cite{shi23} can be added and in a more elaborate setting even translated into skills \cite{wan23voy} - setting the path to modern continuously learning agents that update their own instructions and knowledge \cite{opcl26sel}.



\begin{table*}[h!]
\centering
\footnotesize
\setlength{\tabcolsep}{3pt}
\begin{tabular}{|p{2cm}|p{3.8cm}|p{5.8cm}|}\hline
\textbf{Aspect} & \textbf{Generative AI} & \textbf{Agentic AI} \\ \hline  
Reasoning &  \ding{55} immediate responses & \ding{51} iterative planning and reflection\\  \hline
Interaction &  \ding{55} (mostly with user) & \ding{51} user, tools, real-world, other AI agents \\ \hline
Execution capability  & \ding{55} single-step tasks & \ding{51} Workflows, sequence of actions requiring diverse expertise \\   \hline  
Adaptability &  \ding{55} no self-improvement, bound to training data & \ding{51} Collecting and leveraging experiences \\ \hline 
Autonomy &  \ding{55} user-driven & \ding{51} self-directed, pursuing goals \\ \hline
\end{tabular}
\caption{Comparison of Generative AI vs Agentic AI characteristics} \label{tab:compCap}
\end{table*}

\begin{table*}[h!]
\centering
\setlength{\tabcolsep}{3pt}
\footnotesize
\begin{tabular}{|p{2.8cm}|p{3.5cm}|p{5.3cm}|}
\hline
Autonomy Level & Paradigm & Description \\
\hline
0 - No Autonomy & Classical Machine Learning & 
Tackles narrow tasks it was explicitly trained for. \\
\hline
1 - Assistive Autonomy & Generative AI & 
Handles simple tasks with direct instruction. \\ 
\hline
2 - Partial Autonomy & Agentic AI: Agent-Oriented Workflow & 
Manages multi-step tasks with human oversight and intervention.\\ 
\hline
3 - High Autonomy & Agentic AI: Goal-Oriented Collaboration & 
Achieves complex tasks with occasional guidance. \\
\hline
4 - Full Autonomy & Agentic AI:  Autonomous Decision-Making & 
Given goals, handles all aspects of tasks independently. \\
\hline
\end{tabular}
\caption{Autonomy levels for GenAI and Agentic AI based on~\citep{age25ag}}
\label{tab:aut}
\end{table*}

\subsection{Capabilities Compared}
In terms of their capability, Agentic AI differ from early GenAI in two foundational aspects: (i) Reasoning and (ii) interaction with an environment and tools. Increasingly, Agentic AI also shows a third aspect (iii) continual learning. Other characteristics stated in the literature, such as adaptability, execution capability, and autonomy (shown in Table \ref{tab:compCap}), depend heavily on reasoning and interaction. Regarding autonomy, GenAI can be viewed as assistive technology for simple tasks, requiring specific user instructions for each interaction, whereas Agentic AI demonstrates at least partial autonomy by executing multi-step tasks (see Table \ref{tab:aut})~\citep{age25ag}. Agentic AI allows users to define their own agents using less detailed descriptions that may include identity, professional role, constraints and guardrails, tool-use capabilities, etc.~\citep{par23gen,hon23} -- see Figure \ref{fig:crewSample} for a concrete example. However, as of now, autonomy for both GenAI and Agentic AI is constrained by legislation requiring human oversight~\citep{EU2023AIACT}. Also adaptability including autonomous self-improvement and learning from experiences over a long time-span, is difficult as  evidenced  by prominent failures of earlier systems~\citep{bbc16tay,cdo22ble} and remains limited for Agentic AI. While many works focus on ``LLM agents,'' generalist agents possess multi-modal capabilities and can perform a wide range of tasks, including playing Atari games, controlling a real robot arm, and engaging in ordinary conversation~\citep{reed22g}.

\subsection{Why Agentic AI?}
Generative AI comes with a series of shortcomings some of which are shown in Table \ref{tab:compCap}. They motivate further development. More specifically, Agentic AI promises to eliminate or at least reduce some GenAI limitations, as summarized in Table \ref{tab:limgen}. One of the most pressing limitations of GenAI is its \textbf{execution capability}. Current state-of-the-art GenAI models struggle with moderately complex tasks that require multiple actions, such as simple browser interactions~\citep{drou24,zhou23web}, e.g., ordering a product in a webshop. Although early GenAI like GPT-3\cite{bro20} demonstrated basic reasoning\cite{wei22cha} and tool usage capabilities, it remained highly limited—similar to early AI systems like ELIZA in the 1960s, which could chat with humans but in an unsophisticated manner. 

This lack of capability can be partially compensated through sophisticated prompt engineering, such as providing detailed instructions on how to approach a problem or in-context learning requiring samples as part of the prompt \cite{bro20}. However, reasoning models require only high-level goals and can derive detailed solution steps on their own, while solving a task, i.e., at test time, thereby enhancing \textbf{usability}. 

The paradigm of scaling training data, models, and compute led to the breakthrough of foundation models~\citep{bro20,kap20,chu24sca}. Also, adding samples of similar tasks with a solution process to the prompt can enhance performance often called few-show prompting \cite{bro20}. Both approaches may continue to boost AI’s capabilities, but they are not necessarily optimal.  First, there is a significant lack of \textbf{training and few-shot data}. Data can be expensive to collect—or, in some cases, impossible to obtain. It is not feasible to anticipate all potential tasks and gather large amounts of domain-specific training data for each, even a small number can be challenging as it might require experts defining adequate solution strategies. 

Early GenAI models had very limited \textbf{memory}. For example, ChatGPT 3.5 could only handle inputs of about 4000 words without access to external data sources. More modern foundation models part of Agentic AI can handle millions of input tokens and perform sophisticated retrieval of information to stay up-to-date, support reasoning, and reduce errors. Furthermore, an agent consisting of a model embedded in a harness leverages complex mechanisms for memory management such as compacting reasoning traces and tool outputs to reduce context, i.e., input tokens, as demonstrated for Anthropic's agentic coding harness ``Claude Code'' \cite{liu2026dive}. In contrast, early GenAI even if capable to interact multiple rounds in chats lacked mechanisms for sophisticated memory management.

Early GenAI generally lacked the capability to \textbf{learn}. GenAI’s limited context window and lack of memory management allowed it to consider only a few past user interactions, making it incapable of remembering context or feedback in long conversations. Modern models have much larger context windows and can dynamically retrieve relevant information from databases. Agentic AI systems can learn in a more trial-and-error fashion—e.g., by simulating potential outcomes~\citep{yao23tree} or incorporating real-time feedback from users or tools, such as compiler error messages for generated code. Based on past agent trajectories containing an agent's own actions, reasoning and environment observations, agents can create and adapt themselves through adding such experiences as part of their input \cite{shi23,bro20}. Specifically, agents can adjust their own instructions and create and update skills, e.g., task-specific instructions paired with resources such as software code, which are used dynamically depending on the current problem \cite{opcl26sel,wan26age, wan23voy}. This allows them, for example,  to accommodate for altered or newly discovered user preferences or environmental changes as witnessed by personal assistants such as OpenClaw \cite{opcl26sel}. Still, learning does not encompass continuous changes to the underlying parameters of a model as this has been shown to be unstable more than a decade ago \cite{bbc16tay} and remains an open challenge. Thus, also for agents life-long learning is mostly limited to altering model instructions rather than modifying the underlying model itself. 

Even for tasks that GenAI handles well, Agentic AI’s reasoning capabilities can \textbf{reduce errors} such as hallucinations — i.e., generated content that is unfaithful to the input~\citep{may20}. These may either contradict the source content (intrinsic hallucinations) or be unverifiable (extrinsic hallucinations). Agents that perform self-verification and retrieve external knowledge tend to exhibit lower error rates~\citep{lew20, dzi21}. 

Additionally, academic literature has extensively discussed other GenAI shortcomings, such as \textbf{biases and interpretability}, which remain unresolved despite significant efforts by both industry and academia~\citep{sch24expl,bbc25kle}. Agentic AI improves transparency and interpretability by providing intermediate results allowing for easier verification and better understanding. Furthermore, Agentic AI’s reasoning capabilities offer the potential to reduce biases.

Agentic AI enables \textbf{dynamic and flexible dedication of resources} during inference. For example, to improve performance for a specific task one might invest more computation as shown in Figure \ref{fig:arc}, changing reasoning effort from ``None'' to ``high'' gave more than a 35\% gain in performance. GenAI can only approximate this behavior coarsely by switching between models of different sizes, such as small and large variants. In contrast, an agent encapsulating a single model allows fine-grained control of computation based on intermediate outputs. When resources like computation or electricity are limited, reasoning time—and thus response quality—can be reduced to serve more requests. Computational resources are spent both during model training and later during inference, i.e., when handling user tasks. 

\begin{figure}[h]
\centering
\includegraphics[width=0.65\textwidth]{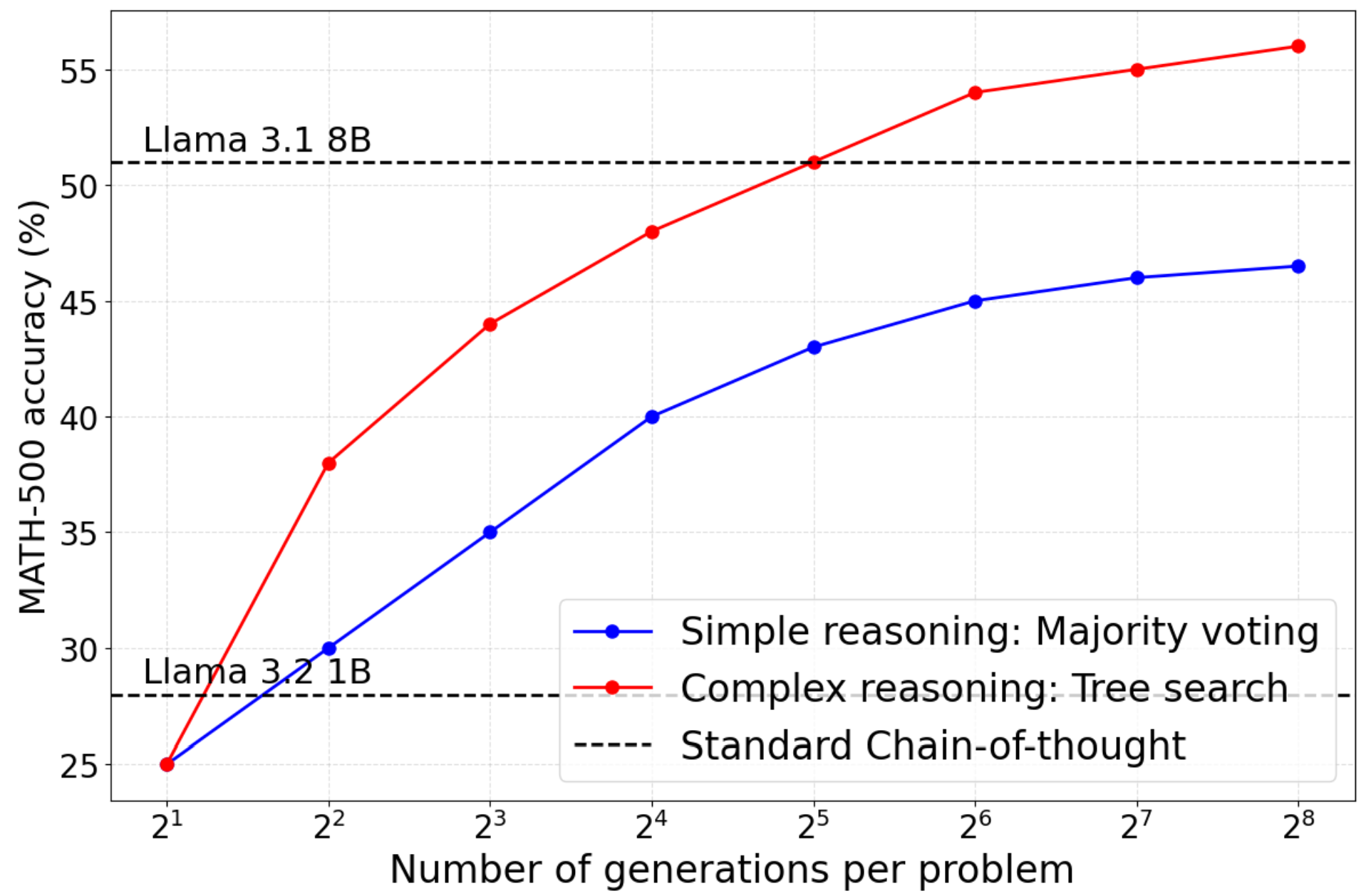}
\caption{A small model with 1 billion parameters (1B) can outperform a larger 8B model in accuracy as shown for the MATH-500 benchmark using more computation, i.e., generations meaning calls to the model. Different ``reasoning'' strategies strongly impact the effectiveness of generations. Figure based on~\citep{bee24tt}. }\label{fig:sca}
\end{figure}

Agentic AI enables novel configurations for improving \textbf{cost-efficiency and amortization of AI products}. Smaller, faster, and less costly models can achieve performance comparable to large models by incorporating reasoning. This makes it economically viable to develop specialized models that are infrequently used, as they require less amortization. Similar to automation in manufacturing, one can choose between high startup costs with low operating costs and limited flexibility (large models), or low startup costs with dynamic, though potentially higher, operating costs (small models with reasoning). Figure \ref{fig:sca} illustrates the performance of a large model using a single generation (i.e., one prompt execution) versus a small model with reasoning that produces a variable number of generations to solve self-generated sub-tasks~\citep{bee24tt}.


\begin{table*}[h!]
\centering
\setlength{\tabcolsep}{3pt}
\footnotesize
\begin{tabular}{|p{3cm}|p{4.6cm}|p{4.8cm}|}
\hline
\textbf{Aspect} & \textbf{GenAI Limitation} & \textbf{Agentic AI Advantage} \\  
\hline
Execution capability & \ding{55} Failing complex multi-step tasks. Limited to generation of digital content with limited tool usage. & \ding{108} Performs multiple steps to solve tasks using planning, interaction with an arbitrary environment and tools. \\ \hline
Usability & \ding{55} Requiring (rather) detailed task execution instructions & \ding{108} Goals without detailed instructions are sufficient. \\ \hline
Training and few-shot data & \ding{55} Relies often on task-specific data, which may be infeasible or expensive. & \ding{108} Leverages logic and external tools to operate with less data.\\
\hline
Learning \& Memory & \ding{55} Limited by small context-windows & \ding{51} Trial and error learning  through simulation and real-world interactions; Larger context memory with external storage paired with ability to change own instructions enable learning from past experiences\\
\hline
Errors & \ding{108} Errors such as hallucinations are common & \ding{108} Less errors due to extensive reasoning and validation at test time \\
\hline
Transparency \& interpretability & \ding{55} Hard to interpret & \ding{108}  Shows intermediate reasoning, making outcomes explainable, though still hard to interpret. \\
\hline
Dynamic and flexible dedication of resources during operation & \ding{55} Little control over computational resources per task & \ding{51} High control allowing to perform more or less reasoning impacting solution quality and costs. \\
\hline
 Cost-efficiency and amortization of AI products & \ding{55} High upfront training costs for large models requiring frequent usage for amortization.
& \ding{51} Low up-front costs due to smaller models with reasoning, supporting economically viable deployment even for infrequent tasks. \\ \hline
\end{tabular}
\caption{Limitations of GenAI (partially) overcome by Agentic AI; \ding{108} indicates areas where further improvements are needed} \label{tab:limgen}
\end{table*}

\begin{figure}
    \centering
    \includegraphics[width=\linewidth]{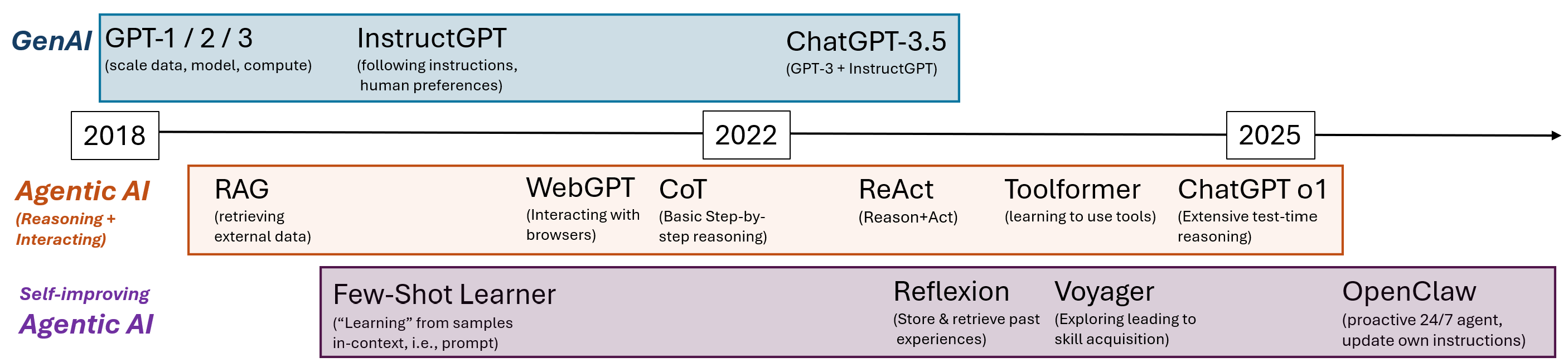}
    \caption{A few milestones from Generative to Agentic AI. Self-improving AI agents are increasingly emerging.}
    \label{fig:genToAg}
\end{figure}

\subsection{Evolution of GenAI to (self-improving) Agentic AI}
Key capabilities needed for GenAI, i.e., general world knowledge and instruction following, and Agentic AI, i.e., related to reasoning, acting, and self-improving, did not evolve sequentially. They can be seen as parallel streams that increasingly entangled - see Figure \ref{fig:genToAg}. 

From a historical perspective focusing on 2019 to 2022, GPT-2~\citep{rad19c}, GPT-3~\citep{bro20}, and the public release of ChatGPT 3.5~\citep{op22cha} --which resulted from post-training GPT-3 for human alignment and instruction following~\citep{ouya22}—marked major breakthroughs in text generation, with substantial differences in performance and task coverage among these models (see top row Figure \ref{fig:genToAg}). In this work, we consider these models as marking the dawn of modern Generative AI, as they enabled prompt-based, controlled text generation of significantly higher quality than before. These ``foundation models'' could address a wide variety of tasks without task-specific training~\citep{schn24f}. Technical foundations such as transformers date back further in time~\citep{sch24com}. Additionally, models like Generative Adversarial Networks (GANs)~\citep{goo14, zha23com} enabled high-quality image generation years earlier, but lacked controllability—especially through natural language instructions. 

\smallskip
\noindent\textbf{Evolution of reasoning:} 
Early generative AI models~\citep{rad19c,bro20,op22cha} typically responded to requests instantly. That is, the input was passed through the model once, without generating intermediate outputs, directly yielding a response—e.g., the prompt “7 $\cdot$ 12 =” produced “84”. However, with suitable prompting~\citep{wei22cha,koj22}, even early models like GPT-3~\citep{bro20} demonstrated basic shallow reasoning, producing longer outputs that detailed the steps taken to reach an answer. These reasoning capabilities became more advanced, as shown in benchmark results (Figure \ref{fig:arc}), due to training on reasoning trace data~\citep{op25gpt} and the application of specialized techniques like planning as well as using verifiers \cite{cob21tra}. Deeper reasoning in Agentic AI systems requires significantly more computation, involving exploration of multiple options (planning and search) that are evaluated and refined through reflection. Scaling computation at inference time allows dynamic control over solution quality, as illustrated in Figure \ref{fig:sca}. 

Several early prompting patterns used in GenAI to enhance reasoning have been integrated into the reasoning processes of Agentic AI. To illustrate this, we highlight a few patterns from~\cite{whi23pro,sch25ag}. The ``alternative approaches pattern'' asking for multiple diverse solutions is found in multiple planning and search methods, such as ToT and FoT. The ``question refinement pattern'' is inherently part of the reflection process in reasoning. The ``cognitive verifier pattern'' follows the idea of problem decomposition~\citep{koj22}, which is a core part of reasoning. The ``recipe pattern'' translates a goal into a sequence of steps, which is also done by agents.

\smallskip
\noindent\textbf{Evolution of interaction, tools, memory, and learning}: 
Early GenAI was already capable of basic interactions with tools like calculators and external memory sources such as databases~\citep{sch23to}. Retrieval-augmented generation (RAG), which uses external data from vector databases, marked a key milestone~\citep{lew20} and was quickly integrated into commercial systems like GPT-4. Tool use and retrieval capabilities have become more advanced in modern Agentic AI systems leading to standards such as the model context protocol (MCP)\citep{ant24mcp}. Furthermore, context window sizes—functioning as short-term, input-dependent memory—have increased from a few thousand to millions of tokens, enabling key Agentic AI features such as instant learning from experience. 

Early GenAI focused on tasks that could be solved with a single generated output and little or no interaction with the environment. Early works targeted interacting with particular software, e.g., using LLMs for browser interaction\cite{nak21web}, but more generic patterns emerged later that combined reasoning and acting emerged later \cite{yao23rea} and learnt tool usage \cite{sch23to}. Whereas GenAI relied essentially only on model capabilities, agents consist of the model embedded in a harness that serves a number of purposes such as memory (context) management and turning model outputs into actions as illustrated by  coding agent harnesses \cite{liu2026dive} and personal agent harnesses \cite{opcl26sel}. Harnesses also ensure that agents can update and change their own instructions allowing them to learn from experiences.
Agentic AI increasingly incorporates the established paradigm of reinforcement learning (RL), “in which an agent interacts with the world and periodically receives rewards that reflect how well it is doing”~\citep{rus21}. RL targets tasks that involve a sequence of actions, each influencing the environment. An agent may continuously sense its environment to monitor for changes due to its own actions or other events. It may periodically receive feedback on its performance. Solving a task may require an agent to make multiple attempts, learning from both failures and successes. Agents also face challenges such as exploitation (using existing knowledge) and exploration (acquiring new knowledge)~\citep{rus21}. That is, exploration aims at getting more knowledge about poorly understood aspects of the environment through novel behaviors and learning from observations to derive better solutions, even though there is very high uncertainty of whether the exploration is successful. In contrast, exploitation uses the agent’s existing knowledge to solve a task, typically yielding only incremental insights. During exploitation a person might take the shortest way to work as every day as it is the fastest known route. During exploration a person might take the subway for the first time.

\smallskip
\noindent\textbf{From narrow prompting to defining autonomous agents}:
Agents rely on customized versions of foundation models with additional functionality encoded in harnesses, e.g., performing orchestration to manage tool access and user interactions. Agents are typically defined via textual descriptions that include roles or personas, workflows, failure, recovery, and permitted tool usage, which implicitly also constrains the action space~\citep{par23gen}; see Figure \ref{fig:crewSample}. To solve a concrete task, an agentic system can simply be prompted as GenAI. The system then engages in a dynamic, stateful solution process, often generating a series of prompts, as seen in early Agentic AI platforms like AutoGPT~\citep{auto23}. However, GenAI also adopted customization of foundation models through prompting. For example, defining roles and personas has gained attention as a key prompting pattern~\citep{wan23rol,whi23pro}. This has become a standard prompting technique in GenAI. By incorporating such definitions into the system prompt~\citep{lee24ali}, users can effectively create customized foundation models. This process has been supported by commercial platforms such as OpenAI’s GPT Store~\citep{gptstore23}, which facilitates easy sharing of models with specific system prompts and potentially private data.
Additionally, the level of abstraction differs when defining an agent versus specifying a prompt in a GenAI context. For example, defining workflows in Agentic AI is arguably more high-level as reasoning models are expected to fill in more detailed, missing steps compared to more low-level instructions in GenAI (as done in early chain-of-thought prompting~\citep{wei22cha}). Moreover, reasoning is now a built-in feature of Agentic AI, whereas in GenAI it had to be elicited using specific prompting patterns~\citep{whi23pro}.

\begin{figure}[h]
\centering
\includegraphics[width=0.55\textwidth]{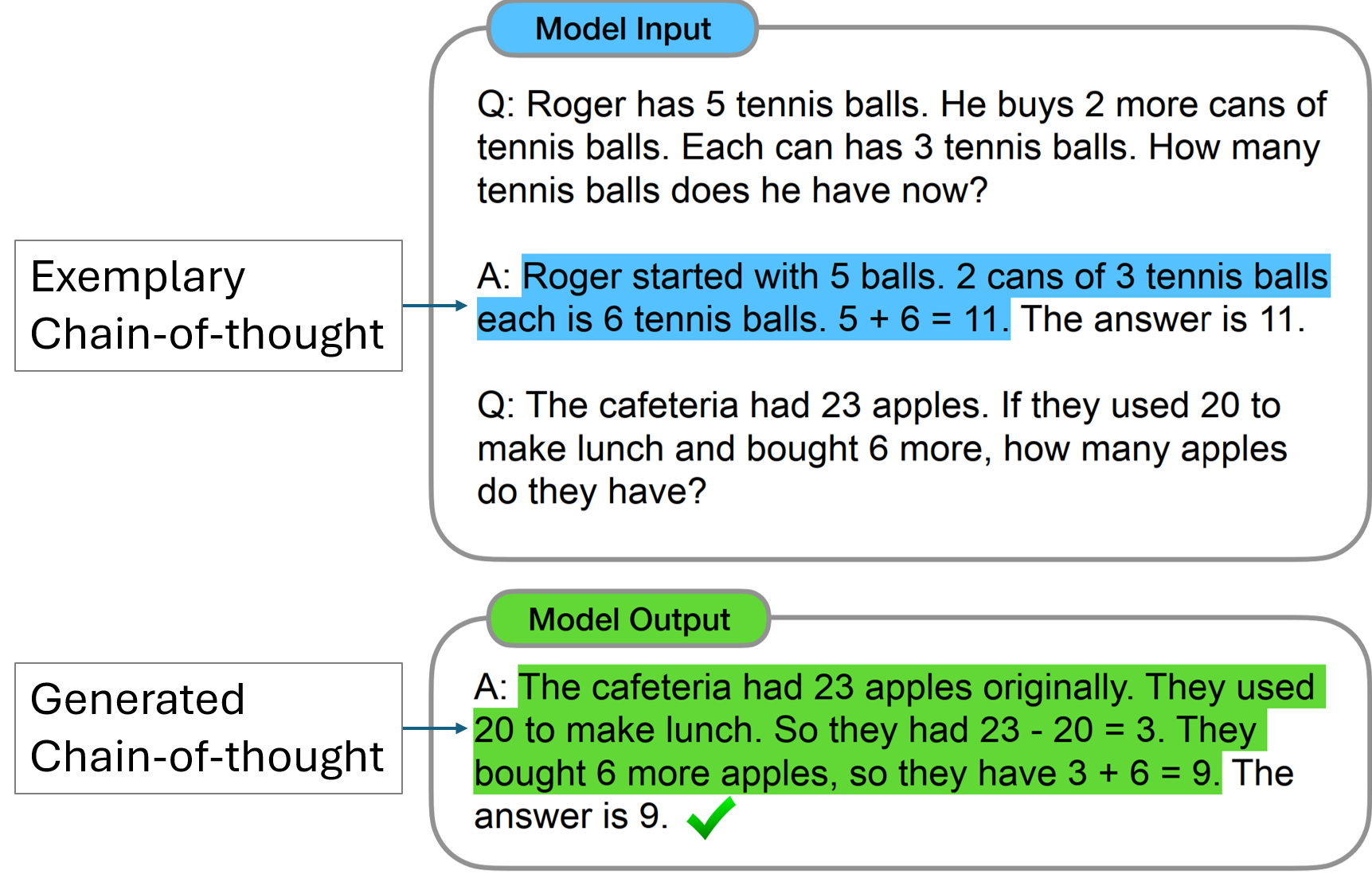}
\caption{Chain-of-thought(CoT) detailing steps to solve a task in the input elicit a CoT in the response thanks to in-context learning (Figure adjusted from~\citep{wei22cha})}\label{fig:cot}
\end{figure}

\section{Agentic AI} 
We elaborate on four essential areas that distinguish Agentic AI and GenAI: reasoning (problem decomposition, verification, search and planning), interaction (with the environment, tools), memory\& learning  and specification of single and multi-agents systems.

\subsection{Reasoning}
Agentic AI performs reasoning. Reasoning can be broadly defined as “the action of thinking about something in a logical, sensible way”~\citep{oxfrea}.  In the AI literature multiple types of reasoning are discussed such as inductive reasoning (generalizing from examples and experiences) and deductive (application of rules). Analogical prompting has also been used as a reasoning paradigm for LLMs~\citep{yas23la}. However, the most prevalent form applied by agents is the creation and execution of a step-by-step solution process~\citep{koj22,wei22cha} at inference involving, potentially, problem analysis, planning, solving, reflection with validation and refinement. This contrasts with an immediate, intuitive memory-based response—analogous to Kahneman’s concepts of fast and slow thinking. In fact, some approaches allow agents to dynamically choose between slow and fast thinking~\citep{lin23swi}. Technically, a reasoning model does not immediately provide the task solution; instead, it first generates intermediate results for self-defined subproblems—e.g., for the prompt “7 $\cdot$  12 =”, it might decompose the task into “2 $\cdot$  7 = 14, 7 $\cdot$  10 = 70, 14 + 70 = 84” rather than directly responding with “84”. Alternatively, it might first decompose the problem into “2 $\cdot$  7 = a, 7 $\cdot$  10 = b, a + b = result” and then solve each subtask to arrive at the final answer using a tool such as a calculator. 

\smallskip
\noindent\textbf{Shallow and deep reasoning:}  
Shallow reasoning appeared in early GenAI systems, typically triggered by prompting~\citep{koj22,wei22cha}. However, evaluation of the reasoning capability of LLMs suggest that they rely on patterns and correlations found in training data rather than on  reasoning abilities~\citep{Mon24bey}. Furthermore, (simple) CoT primarily helps on math and symbolic reasoning~\citep{spra24cot}. Deep reasoning is more extensive, better structured, and pursued automatically in modern Agentic AI systems. It often integrates algorithms—e.g., for planning~\citep{yao23tree} and verification\cite{meni22}—and tool use, where problems are translated into code and solved via a code interpreter~\citep{gao23pa,chen22pro}.

\subsubsection{Decomposition}

\noindent\textbf{Multi-step solutions:} Problems can be solved through multi-step reasoning in various ways—either via a single prompt generating a long answer or through a more intricate process involving multiple agents with separate planning and reflection, as shown in Figure \ref{fig:reasEx}. The latter is the more prevalent paradigm in Agentic AI. Basic decomposition can be achieved through prompting, e.g., by providing a multi-step reasoning example as in Chain-of-Thought (CoT)~\citep{wei22cha} (see Figure \ref{fig:cot}), asking the model to “Think step-by-step”~\citep{koj22}, or explicitly instructing it to decompose and then solve each subtask, as in least-to-most decomposition~\citep{zho22le}. Recent work also showed that decoding strategies can elicit reasoning, in particular, CoT generations have higher confidence~\citep{wan24cha}.

``Planning is defined as the task of finding a sequence of actions to accomplish a goal''~\citep{rus21}. Search is ``the computational process of planning''~\citep{rus21}. For instance, tree search algorithms aim to identify a sequence of actions that leads to a goal. Early large language models typically did not perform explicit search before solving a task but relied on prompting to execute step-by-step reasoning~\citep{wei22cha,koj22}. When designing reasoning demonstrations for prompts, as in in-context learning in the classic Chain-of-Thought paper~\citep{wei22cha}, two key dimensions emerge: (i) content within each reasoning step (e.g., answer accuracy, use of reasoning keywords) and (ii) structure (e.g., reflection, validation, logical coherence). Recent work has shown that structure is much more important~\citep{li25llms}. Moreover, the length of reasoning chains—such as decomposition into smaller problems—is more impactful than the difficulty of individual components~\citep{she25}. Though longer chains often perform better, verbosity is not helpful and concise chains-of-thought can be more effective~\citep{xu25dra}.

\smallskip
\noindent\textbf{Hierarchical decomposition:} While CoT resembles sequential planning, the concept of hierarchical planning has also been explored~\citep{aja23,yan24buf,wu24bey}. In~\citep{aja23}, high-level steps are proposed and then refined into more concrete geometric and control-level actions. Different levels of abstraction use different models—for instance, a large language model (LLM) for high-level planning and a visual model for trajectory generation as a “geometric plan.”~\citep{yan24buf,wu24bey,yan25rea} generate high-level solution guidelines and retrieve them for specific problems. These guidelines are then elaborated upon to solve the target problem.~\cite{yan25rea} created a library of approximately 500 generic thought templates and trained models using them. In contrast,~\cite{wan24st} first derives a high-level solution strategy and then retrieves previously generated demonstrations aligned with that strategy to solve the problem. The idea of first elaborating on task specific reasoning has also been proposed, e.g.,~\cite{zhou24,gao24met}. In~\cite{gao24met}, a model first selects a reasoning method (e.g., CoT, ToT, self-refine) and then solves the problem using that method.

\subsubsection{Reflection} 

\noindent\textbf{Reflection} in the context of learning is ``exploring one's experiences in order to lead to new understanding and appreciations''~\citep{bou85ref}. AI literature has widely discussed two core concepts: (i) verification, which assesses or validates generated outputs such as the quality or truthfulness of step-by-step solutions or partial results, and (ii) refinement, which aims to improve prior outputs. A key question is what information is incorporated in the reflective process, especially regarding the availability of external feedback. A model may reflect on its output without external information (e.g., during planning), or it may use observations and direct feedback from the environment.  Although reflection is considered essential to reasoning, the capacity of LLMs to reflect using simple prompts without external input has been questioned~\citep{hua24la, zha24se}. Some improvements have been attributed to incorporating feedback from oracles, such as compilers~\citep{chen23} or maybe attributed to inherent task structure rendering verification simpler than task solution \cite{cob21tra}. 
Another concern is whether reflection occurs in (continuous) latent space (e.g., \cite{hao24tra,chen24lan}), leveraging internal states prior to being turned into tokens, or discrete tokens, i.e., based on human-readable, generated model outputs.

\smallskip
\noindent\textbf{Verification:} Verification allows to choose the most appropriate plan prior to taking any actions or stop (and restart) the current solution process, if intermediate results are wrong or deemed highly uncertain.
Using multiple alternative chain-of-thoughts for verification—i.e., to ensure self-consistency—has been proposed~\citep{wa22self}. This ensembling method improves output quality through majority voting; for example, if a model produces four answers—4, 8, 7, and 4—it would select “4” as the final result. \cite{cob21tra} trained a small verifier model to assess the correctness of math tasks, showing that choosing results highest ranked by the verifier improves performance.  ~\cite{im23ma} focused on explicitly verifying (i) intermediate rather than final solutions, (ii) through diverse solution strategies, and (iii) using external tools as solvers. Specifically, they reformulated math problems to be solvable by both an algebraic solver and a Python interpreter. They compare the outputs of both approaches. If the outputs were not identical, further invocations were initiated.~\cite{chen23} aimed at code error correction by using debugging samples (in-context learning) and using feedback from a compiler to assess generated code. The approach of assigning confidence scores to individual reasoning steps—rather than to the entire solution—has also proven effective~\citep{raz25cer}. In principle, the same model that is used for generation can also be used for verification, though specialized models trained for verification can lead to better outcomes~\citep{hos24v}.

\smallskip
\noindent\textbf{Refinement:} Progressive-hint prompting improves results iteratively by appending prior outputs to the input~\citep{zhe23}. For instance, if the model initially answers “5,” the next iteration might append “The answer is near 5” to the prompt. In the self-refine method~\citep{mad23}, a separate prompt is generated to collect feedback on the model’s output. The feedback, along with the initial input and output, is used to produce a refined solution. This process is repeated by appending each iteration’s output and feedback to the next prompt. In this way, the LLM builds a “chain of reflections” by accumulating prior outputs. This concept of iterative accumulation appears in other self-reflection studies as well~\citep{shi23,yao23rea}. Furthermore, guiding the reflection process through a meta-reflection process has also been proposed~\citep{liu25ins}.

\begin{figure*}[h]
\centering
\includegraphics[width=\textwidth]{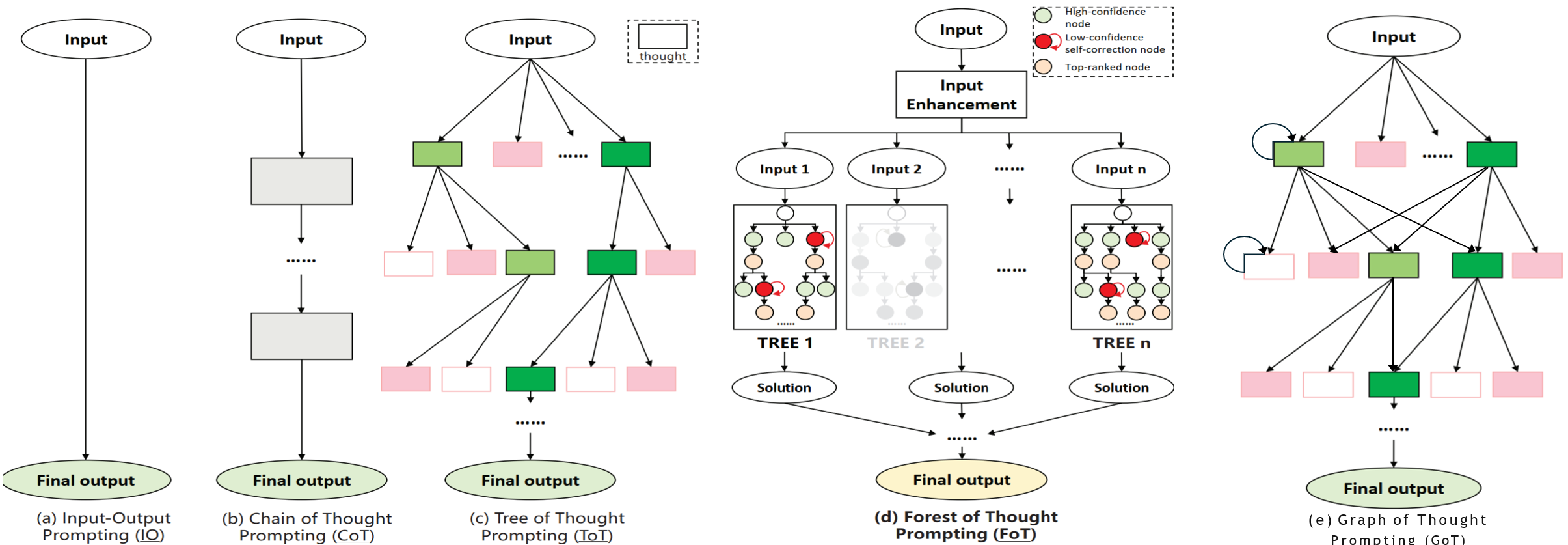}
\caption{Evolution of reasoning from direct input output prompting, to Chain-of-thought to a forest and graph of thought (Figure enhanced with GoT from~\cite{bi24fo})}\label{fig:cotEvo}
\end{figure*}

\subsubsection{Planning and Search}
Tree of Thought (ToT)~\citep{yao23tree}, Forest of Thoughts (FoT)~\citep{bi24fo}, and Graph of Thoughts (GoT)~\citep{bes24} represent further developments of the idea of generating multiple reasoning paths, as in self-consistency~\citep{wa22self}. These paths are assessed and filtered to yield improved outcomes—see Figure \ref{fig:cotEvo}. These approaches share the view that evaluating complete chain-of-thoughts may be computationally expensive for several reasons: It is often preferable to halt solution generation early if the path appears unlikely to produce a good result. If an intermediate step is incorrect, it may be more effective to refine that specific step before proceeding with the reasoning process. Additionally, if a reasoning chain yields strong intermediate results, it may be beneficial to retain the initial steps and explore alternative continuations. Conceptually, after each reasoning step, the partial chain is evaluated to determine whether it is worth pursuing further. Such concepts are well-rooted in the classical search literature. Therefore, extensions of the Chain-of-Thought (CoT) paradigm can be understood as applying classical search techniques in the context of LLMs. For instance, ToT explores strategies like depth-first search—evaluating each chain fully—and breadth-first search, where multiple partial chains are expanded concurrently by choosing the next step for each, potentially growing the set of active chains. Pruning is employed to terminate the progression of partial chains that receive poor evaluations. 

\begin{figure}[h]
\centering
\includegraphics[width=0.45\textwidth]{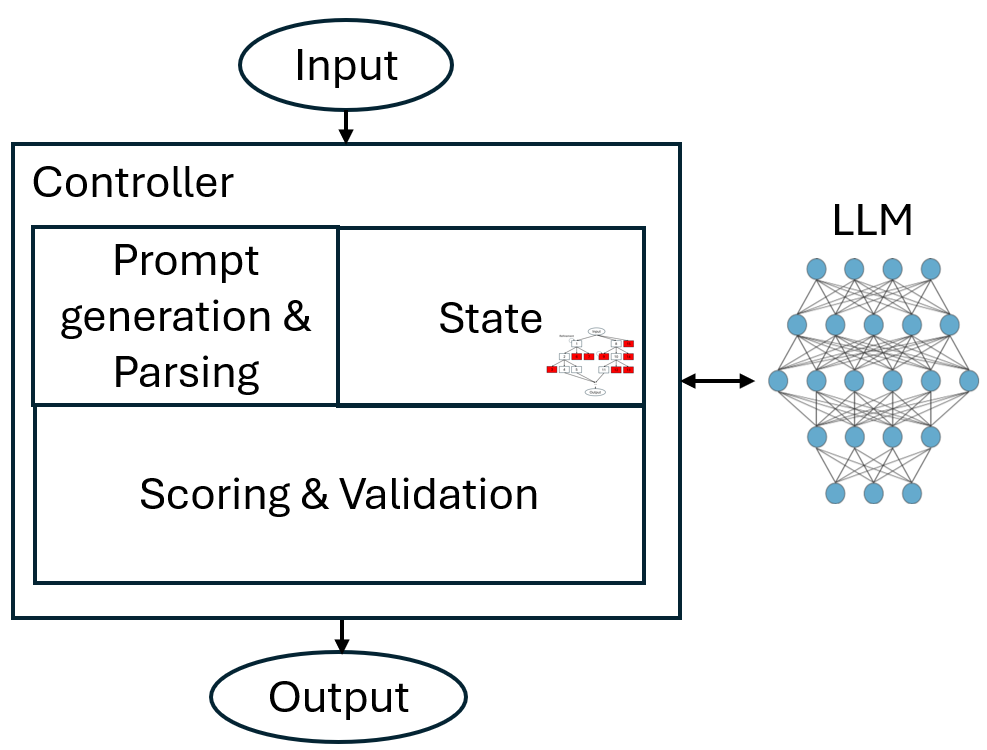}
\caption{Architecture for CoT combined with search with a separate controller containing also Scoring and validation.}\label{fig:cotsea}
\end{figure}

\smallskip
\noindent\textbf{Assessing intermediate results:}
A major challenge lies in evaluating partial solutions when neither the final outcome nor the correctness of intermediate steps is known. To address this, methods like ToT use self-consistency voting~\citep{wa22self} or enlist an LLM as a solution evaluator~\citep{bes24,bi24fo}. Human evaluators~\citep{bi24fo} and problem-specific heuristics can also serve in this assessment role. That is, though ideally, general methods are used, task-specific assessment methods can be more effective.
As a result, such systems typically include components beyond the LLM. A controller maintains the search state, generates prompts, parses outputs, and ranks candidate next steps. While scoring and validation may be handled by the controller, they are often performed by an LLM—see the common architecture in Figure \ref{fig:cotsea}. In the general graph framework~\citep{bes24}, specific prompts are used to generate subtasks (i.e., graph nodes as solution steps), solve them, assign scores, evaluate them, and merge the resulting thoughts. Alternatively, a model can be fine-tuned on search data to enhance its search capabilities~\citep{gan24} or to internalize the search process itself~\citep{schu24mas}, potentially obviating the need for external search algorithms.

\smallskip
\noindent\textbf{Diversity:} FoT and other approaches~\citep{bee24tt,lin25en} aim at diversity. FoT generates variations of inputs, so that for each input a separate tree is generated. In~\citep{lin25en} multiple agents are responsible for diverse thoughts, which are also stored (if successful) and retrieved~\citep{lin25en}.~\cite{bee24tt} uses beam search paired with a (domain-specific) process reward models to achieve diversity.

\smallskip
\noindent\textbf{Planning:} Similar to search, existing planning techniques and tools have been adopted~\citep{liu23llm,hao23rea}.~\cite{liu23llm} generates as output instructions that are executed by a specific planning tool. In contrast,~\cite{hao23rea} adopts Monte Carlo Tree Search (MCTS), where the LLM incrementally builds a search tree. LLMs have also been used with heuristic planning, where an LLM generates actions, evaluate their feasibility and long-term payoff leveraging learnable domain knowledge~\citep{haz24}.  For more information on how LLMs are incorporated in planning consult the survey~\citep{pal24}.


\subsubsection{Learning to Reason}

\noindent\textbf{Pretraining} using next-word prediction has remained largely consistent since the early days of LLMs~\citep{rad19c,bro20, touv23,met25lla}, with autoregressive decoder-only models continuing to dominate. Multi-modal models leverage distinct encoders to deal with diverse data such as text and images~\citep{tou24}. Encoder-decoder models like T5-Flan~\citep{chu24sca} or bidirectional models like BERT~\citep{dev19} have become less prevalent, though are still heavily researched, e.g.,~\cite{sch24imp} uses a bidirectional model for verification of single tokens —generating next tokens via a decoder and predicting the second-last token using a verifier model. 

\smallskip

\noindent\textbf{Training to reason: } The reasoning capabilities of LLMs, often elicited through prompts like “Think step-by-step”~\citep{koj22}, are instilled during training using datasets that include examples of step-wise reasoning. In other words, training models with next-word prediction on data containing step-wise reasoning~\citep{bro20} can endow them with basic reasoning skills, but reasoning can also be enhanced using reinforcement learning as part of a post-training phase  \cite{yan25qw}. Therefore, an obvious next step for enhancing reasoning is to gather more reasoning-focused data and train on it. While the original CoT paradigm~\citep{wei22cha} required users to provide example reasoning paths, since then it has been proposed to incrementally generate CoTs and alternate between their generation and training~\citep{zel22sta}. This approach requires a small number of demonstrations and a larger database of tasks and solutions (but without any rationale). Specifically, the LLM generates a new CoT for a given question by leveraging related existing demonstrations. If the generated solution is correct, its CoT is added to the demonstration dataset. The model is then fine-tuned on this newly expanded set of CoT demonstrations. Training on domain-specific data also enables domain-specific reasoning, as shown for the medical domain~\citep{zha023hua}.
While reasoning models excel in task performance, they also suffer from overthinking\cite{che24donot}, i.e., spending an unnecessary large amount of computation for very simple problems. This inefficiency problem can be tackled using model oriented approaches (e.g., rewarding shorter reasoning chains), output oriented approaches (e.g., compressing reasoning steps), and input-prompt oriented approaches (e.g., specifying output limits)\cite{sui25stop}. Recently, so-called hybrid reasoning models like ChatGPT-5 and QWEN-3 have gained in popularity, where a user (or the model itself) deliberately decides on how much computation should be spent for a prompt.

\smallskip
\noindent\textbf{Training a state-of-the-art hybrid model: QWEN-3.} The QWEN-3~\cite{yan25qw} foundation models were first trained using next token prediction on about 30 trillion tokens, followed by a reasoning stage consisting of 5 trillion tokens containing an increased proportion of coding, reasoning, STEM, and synthetic data. To further enhance reasoning as part of post-training, reinforcement learning was used using about 4k query-verifier pair, e.g., for math a query might be ``Compute derivative of $\sin(x)+x^3$'' with a symbolic math engine (such as sympy) to verify if the model output is correct. Furthermore,  a so-called thinking mode fusion training phase based on supervised fine-tuning was employed to learn to control the amount of computation for a prompt based on whether the user wants the model to think (for longer) or to obtain an instant answer.

\smallskip
\noindent\textbf{Training a state-of-the-art (reasoning) model: DeepSeek.} Although reasoning models generally follow the same pre-training phase, subsequent training stages may differ. After pre-training, models are typically subjected to supervised fine-tuning (SFT) using high-quality instruction datasets. Without this step, later (reinforcement learning-based) methods might struggle to meaningfully improve reasoning due to a poorly grounded base behavior.
Accordingly, the DeepSeek paper~\citep{guo25} offers a detailed overview of a recent state-of-the-art reasoning model called DeepSeek-R1 and the earlier works that contributed to its development.\footnote{OpenAI's training process for o1 and o3 is not public.} 

DeepSeekMath~\citep{sha24dee} carefully curated web-data to identify data relevant for mathematical reasoning. This data was used to train models via a reinforcement learning technique known as Group Relative Policy Optimization (GRPO). GRPO evaluates behavioral policies by averaging output groups instead of computing explicit value functions or models, which are more computationally intensive. That is, say an LLM produces hundreds of (similar) outputs for (similar) prompts. The outputs are grouped, e.g., based on content similarity, and performance measures are computed for the group as an average. Models are optimized towards generating outputs similar to the best performing groups. This method constituted one component in the training process of DeepSeek-R1~\citep{guo25}. The best performing DeepSeek model variant built upon a strong base model (DeepSeek V3). First, it was trained on a small set of high-quality CoT examples. Without this data, the model’s outputs were significantly less human-readable. Second, it applied GRPO reinforcement learning using a reward function that considered: (i) solution accuracy—assessed via rules or external tools like code interpreters; (ii) formatting, to ensure output usability; and (iii) language consistency—avoiding mixed-language responses (e.g., Chinese and English), though this slightly reduced performance. The third step involved supervised fine-tuning on a mix of self-generated reasoning and non-reasoning data, along with the data used to train DeepSeek V3. Finally, to achieve human alignment (helpfulness and harmfulness) another training stage is conducted.

\smallskip
\noindent\textbf{Incorporating human preferences:} LLMs have also been trained to reflect human preferences—rather than step-wise reasoning instructions—using RL-based methods and supervised training, such as training on generated sentences conditioned by user feedback~\citep{raf23,zie19,liu23cha}. Typically, humans rate two or more reasoning options, and the LLM learns to infer latent factors that capture these preferences. However, to reduce data collection costs, LLM-generated feedback has been used to emulate human responses~\citep{dub23alp}. Nevertheless, emulated feedback must be high-quality, as data quality is considered more critical than algorithm choice for achieving strong performance~\citep{ivi24}. While these methods are vital for improving human-LLM alignment (in terms of helpfulness and safety), they are less commonly used to enhance reasoning, though they can be applied for that purpose. When oracle feedback for the final output is available, it can be used to derive rewards for intermediate steps~\citep{zha24re}, enabling the creation of training data for self-supervised learning and enhanced reasoning.

\begin{figure*}[h]
\centering
\includegraphics[width=0.8\textwidth]{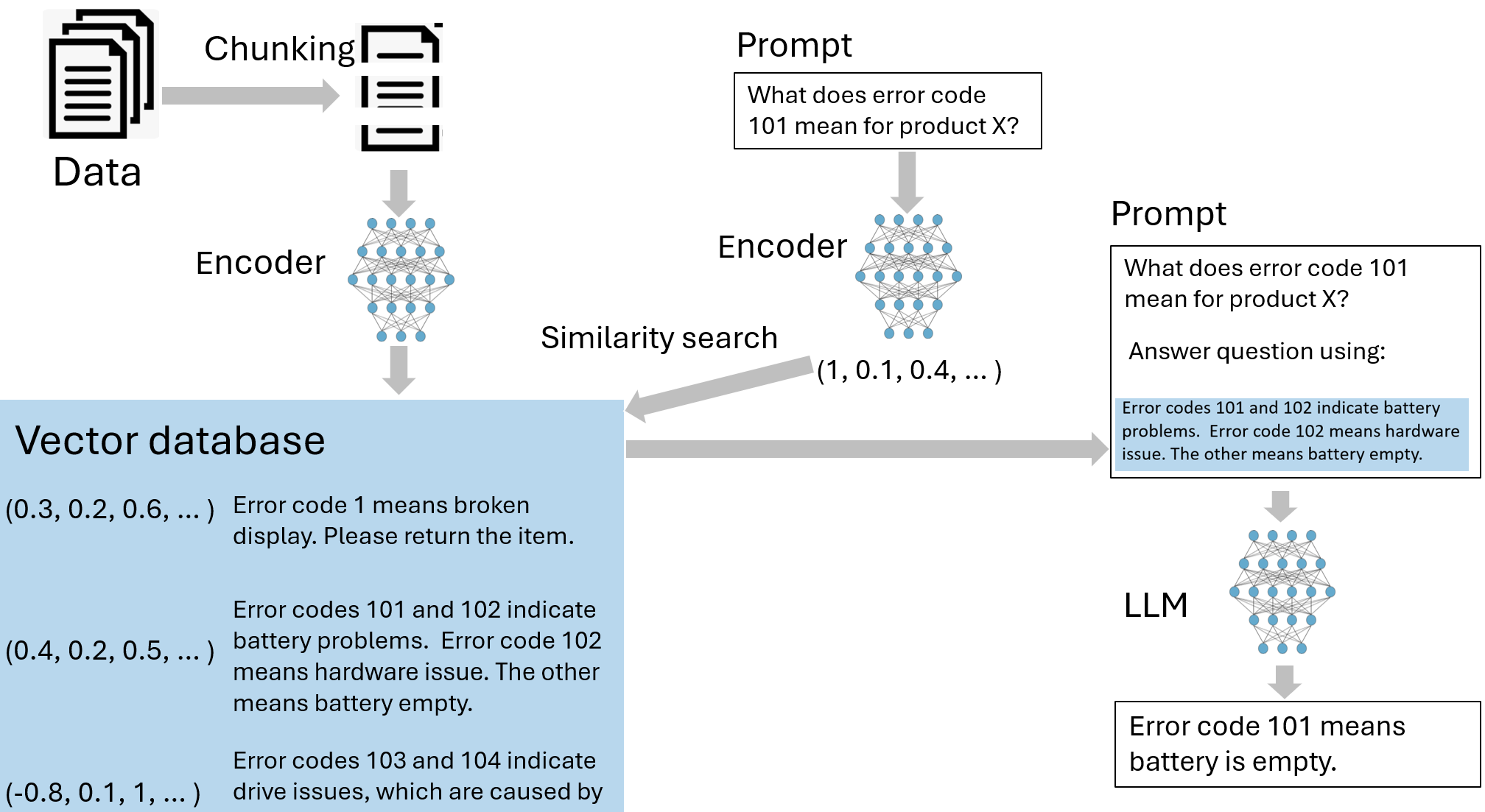}
\caption{Retrieval augmented generation (RAG)~\citep{lew20} enhances prompts with external information. It requires a vector database where  vectors summarize text chunks and serve as index for retrieval. }\label{fig:rag}
\end{figure*}

\subsection{Memory \& Learning } 
       
\begin{table}[h]
\centering
\footnotesize
\setlength{\tabcolsep}{3pt}
\begin{tabular}{|p{2.3cm}|p{4.7cm}|p{4.7cm}|}\hline
\textbf{Characteristic} & \textbf{} & \textbf{} \\
\hline
Memory functions  & \multicolumn{2}{p{9.6cm}|}{Reasoning, personalization, processing large data and learning} \\ \hline
Memory Persistence & Short-term (Working memory, model context) & Long-term (parameters, databases) \\ \hline
Memory Type & \multicolumn{2}{p{9.6cm}|}{Architectural, retrieval-based, parametric, and ephemeral memory} \\ \hline
Memory Location & Internal (parameters, context window) & External (databases, APIs, accumulation of inputs/outputs), and hybrid\\ \hline
Information Source & Agent-discovered (feedback, reflection), Agent-generated (simulation) & External, non-agent knowledge (general world knowledge) \\ \hline
Memory access & Non-retrieval (direct access of relevant information) & Retrieval-based \\ \hline
\end{tabular} 
\caption{Characteristics of Memory and Stored Information} \label{tab:mem}
\end{table}

\subsubsection{Memory Types and Structures} 
Memory and information characteristics are summarized in Table \ref{tab:mem}.
\textbf{Memory functions} are multifaceted. Memory is essential for processing user prompts—early models were unable to handle large prompts due to small context windows. This limitation hindered applications such as personalization and the processing of large input data, e.g., books. Memory is also relevant for reasoning as reasoning commonly requires solving problems in sub-steps, which leads to more tokens being generated than just outputting a solution. Furthermore, staying up-to-date (e.g., learning about changes and novel developments after model training, i.e., after collection of the training data) needs either updating the model parameters or access to memory containing such (newer) information. Similarly, to improve on failures requires that novel experiences are leveraged, which must be stored somewhere. 

Technically, model parameters represent a form of long-term \textbf{persistent memory}, encoding training data via a learning process involving fitting and compression, as in typical deep learning models~\citep{sch24un}. Although fine-tuning model parameters is common, particularly during training, online updates pose risks such as catastrophic forgetting~\citep{fre99ca,kir17ov} or malicious manipulation, as seen in real-world cases in the more distant past (such as the Tay chatbot~\citep{bbc16tay}) but also more recently (e.g., BlenderBot 3 \citep{cdo22ble}). Therefore, model parameters (i.e., parametric memory) are generally treated as fixed in scenarios requiring instant learning through environmental interaction.

Parametric memory represents just one \textbf{type of memory}. Ephemeral memory refers to non-persistent memory used during request processing. It determines how many tokens a model can process simultaneously. Like human working memory, which is limited to about seven items~\citep{mil56}, model context windows are bounded—though modern models can now hold millions of tokens~\citep{tea24gem,met25lla}. In contrast, retrieval-based memory~\citep{lew20} offers virtually unlimited capacity though only limited amounts of retrieved information can be added to inputs (due to limits of the ephemeral memory). However, identifying the ``right'' information is challenging — retrieving relevant information without including irrelevant content is non-trivial. Unfortunately, LLMs are easily distracted by irrelevant content\citep{shi23la}. Architectural memory refers to memory mechanisms built directly into the model architecture. For instance, MemGPT~\citep{pac23mem} introduces a memory hierarchy that mitigates context window limitations by loading from long-term memory and discarding temporarily unneeded information.

\emph{Memory location} can be internal (parametric or ephemeral), external (retrieval-based), or hybrid (architectural memory). In some cases, memory consists solely of the current prompt as well as past inputs and outputs. For example, in a chat, typically information on past utterances of both the user and the model are accumulated and included throughout the chat to any user prompt. In a reflective, iterative process, a prompt might be enhanced in each iteration using past generated feedback~\citep{mad23}. In these cases, memory is continuously accumulated and fed to the model in full, rather than being selectively retrieved based on queries.

\emph{Information source} can be external world knowledge as found in the internet or originate from the agent either from discovery using feedback and reflection on actual task execution or from simulation and, in principle, pure reasoning on tasks not actually executed, e.g.,  training tasks can be solved by an agent and the solution workflow stored for later retrieval to support solving similar tasks \cite{wan24age} and the helpfulness of generated data has been demonstrated also in self-training \cite{hos24v}. 

\subsubsection{Learning from memory}
Learning from memory by adjusting a model's context is a part of agentic context engineering\cite{zha26ace}, where context is maintained through reasoning, retrieval and compression. 
Past experiences can selectively be retrieved \cite{shi23} as over time they cannot be maintained in context due to size and it might also not be ideal to maintain all of them in memory.  Experiences can also be condensed into skills that are acquired and retrieved during agent operation as part of environment exploration \cite{wan23voy}. Memory compression can also be learnt, e.g., by maintaining information relevant to past failures \citep{kan26acon,yan26memr1}. 
This cumulated in more recent, well-known systems such as the personal AI agent OpenClaw \cite{opcl26sel}, which operates continuously and learns from its experiences by updating its own knowledge base and instructions.

The Darwin Gödel Machine (DGM)\citep{zha25dgm} modifies its own code. It grows multiple coding agents, samples them to create new, interesting versions of themselves validated using coding benchmarks. In this way, it also improves its ability to modify
its own codebase.

\subsubsection{Retrieval Augmentation} 
Retrieval augmentation involves selecting a smaller, relevant subset from a large knowledge source to include in the model's input.
Knowledge retrieval reduces hallucinations, incorporates post-training or private data, and enables dynamic learning from experience. Classical RAG approaches~\citep{lew20} use vector databases built by chunking large documents into smaller text snippets. Each snippet is encoded into a vector, which serves as an index for retrieval. During retrieval, the prompt is transformed into a vector to locate relevant text snippets, which are then appended to the prompt, as shown in Figure \ref{fig:rag}. Key design challenges include chunk size, vector encoding strategies, and the number of chunks to include. Several studies aim to address these challenges. For instance,~\cite{asa23} proposes adaptive retrieval through iterative self-reflection. Fine-tuned LLMs can emit retrieval tokens to signal the need for more information.~\cite{zha24a} fine-tunes a model to filter out irrelevant retrieved content.~\cite{yan24} employs an evaluator to classify retrieved knowledge as correct, incorrect, or ambiguous. Incorrect knowledge prompts further retrieval, while correct knowledge is decomposed, filtered, and recomposed before being passed to the LLM. Vector databases can hold external information but can also be used to store and retrieve experiences of an agent interacting with an environment, e.g.,~\cite{zha24ex} stores experiences but also further processes them using an LLM to learn more abstract insights. Experiences may include irrelevant details that hinder retrieval efficiency and unnecessarily occupy the context window~\citep{zhe23syn}.

MemGPT~\citep{pac23mem} proposes a memory hierarchy, which allows to overcome the limited context window size by loading from long-term memory and discarding information not temporarily needed. This architecture is inspired by memory management techniques used in operating systems. Furthermore, there is also work that deals with GPU memory management to improve throughput, i.e., by maintaining past computations (key-value pairs) during the next token generation process~\citep{kwo2023ef}. The integration of search into the reasoning process with GRPO and without labeled data has been successfully performed~\citep{che25re}. 
  
Models with large context sizes often allow to forego retrieving subsets of the data in favor of including the entire information as part of the input. Relying on context only has been shown to lead to better performance at the price of higher generation costs and trade-offs based on dynamic decisions making have been proposed~\citep{li24re}. 

\smallskip
\noindent\textbf{Retrieval from classical databases and knowledge graphs } Aside from vector databases, LLMs can also interact with classical relational databases by generating SQL queries~\citep{li23can}. Furthermore, classical word based indexing methods like BM25~\citep{kar20d} can be used as well, which are good for exact matches (e.g., of rare words), speed, interpretability, and simplicity. Rather than retrieving from databases the idea of retrieving from knowledge graphs has gained a lot of traction as surveyed in~\citep{zha25}. Graphs allow to better decompose and interconnect information. For example, graphs allow to easily represent hierarchies of knowledge. Moreover, several works have developed specialized agents for RAG tasks~\citep{sin25}, such as agents tailored for SQL databases and real-time web data retrieval.

\subsection{Tools}
Tools are external functionalities that can be invoked by an agent. They complement agents and offer several benefits: (i) Reducing errors and increasing efficiency—tools can improve accuracy and speed, such as using a calculator for basic math operations. (ii) Functionality and interaction—tools enable new capabilities for task-solving and engagement; for instance, a web browser allows an agent to perform tasks like online shopping. (iii) Interpretability and control—tools built using classical software can embed rules in a clear and reliable way, enhancing transparency and oversight.

\subsubsection{Tool Creation} Since LLMs can generate code, they are also capable of creating their own tools. The concept of formal tool creation by LLMs has been explored in several works~\citep{wol25, yua23cra, cai23}.~\cite{wol25} converts academic papers containing code into tools accessible by LLMs, while~\citep{cai23,din25too} generates Python utility functions as tools. In addition to generating tools with GPT-4, the process in~\citep{cai23} includes proposing tools using three training examples, verifying them through unit tests, and wrapping them with documentation and examples. Similarly,~\cite{auto23} supports dynamic creation and execution of scripts by agents.

\subsubsection{Tool Selection} 
An agent might be able to handle multiple tools. Different tools support diverse solution strategies. As a result, selecting the appropriate tool becomes an integral part of the planning process~\citep{rua23}. Tool selection can be addressed by adapting the chain-of-thought paradigm~\citep{che23chat}.~\cite{sch23to} employed self-supervised learning to determine optimal tool invocation times. ReAct-style prompting~\citep{yao23rea} integrates reasoning and action, encompassing tool selection as part of the agent's decision-making.

\subsubsection{Tool Use}
~\cite{sch23to, par22,nak21web} utilize software tools to carry out tasks. To this end, they fine-tune an LLM to generate API calls to invoke software and parse its responses covering browsers and simple tools such as a calculator, translation system and search engine. Subsequent research has focused on enhancing fine-tuning and prompting to improve API call accuracy~\citep{pat24}, including support for “out-of-distribution” APIs not explicitly optimized during training~\citep{qin23}. Self-play approaches have been proposed~\citep{par22}, where a small set of tool usage examples is iteratively expanded by adding sampled usages that yield reasonably good outputs. The use of programming language interpreters has become increasingly common. The LLM can output executable code that is run by a program interpreter to answer a prompt. Some approaches~\citep{wa24ex, tas23} implement reasoning through code, refining it step-by-step in response to environmental feedback. LLMs are also capable of generating SQL queries to handle analytical tasks using relational databases~\citep{li23can}. LLMs themselves may be treated as tools, as shown in~\citep{she23}, where tasks are decomposed and subtasks routed to specialized LLMs. Standardization of tool definitions like MCP \citep{ant24mcp} facilitates tool use across model families. 

\subsection{Interacting} 

\subsubsection{Reinforcement Learning} 
The concept of interaction and agents has long been central to AI, as reflected in the definition of agents:  “[Agents] receive percepts from the environment and perform actions”~\citep{rus21}. In reinforcement learning (RL), agents receive rewards—sometimes in the form of human feedback—that signal the quality of their behavior~\citep{rus21}, as illustrated in Figure \ref{fig:gentoAge}. Agentic AI incorporates RL by leveraging foundation models to apply extensive world knowledge. For instance, Voyager~\citep{wan23voy} demonstrates continual learning in Minecraft through exploration, building a skill library, and iterative prompting that incorporates feedback, execution errors, and self-verification. 

\smallskip
\noindent\textbf{World model and policy:} An RL agent simultaneously learns a world model and a behavioral policy. The world model defines possible environmental states, the agent’s state, and transitions between states. The policy dictates which action an agent should take in a given state. Both the world model and the policy are learned from past experiences and received feedback.

\smallskip
\noindent\textbf{Exploration and Exploitation:} Unlike classical supervised or unsupervised learning where data is given, RL agents actively influence data collection through novel actions and observations—i.e., exploration. Agents must balance exploration (acquiring knowledge) with exploitation (applying existing knowledge to perform tasks). RL agents typically execute a sequence of actions to reach goals, each involving perception, reasoning, and action.

\smallskip
\noindent\textbf{RL and Agentic AI:} There is no one-to-one correspondence between RL agents and agentic AI systems. Agentic AI targets broader, more open-ended tasks and may define its own objectives. An RL agent aims to optimize a (more narrow) reward function by learning a policy for how to act in a specific environment. Nonetheless, Agentic AI incorporates many RL concepts, especially during training~\citep{guo25}. RL architectures like actor-critic models—which separate decision-making (actor) from evaluation (critic), often via self-assessment—are also commonly adopted in Agentic AI systems.

\begin{figure*}[h]
\centering
\includegraphics[width=\textwidth]{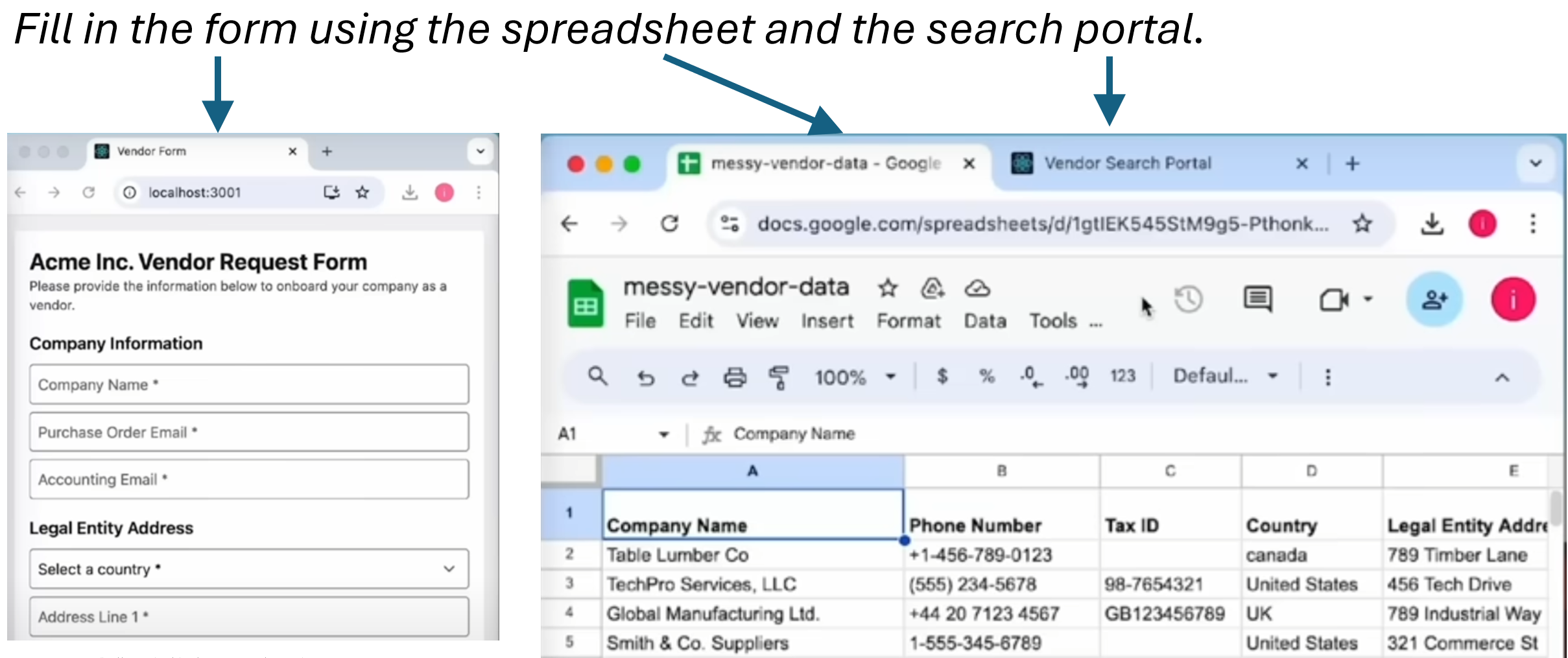}
\caption{Instruction to (Anthropic's) agent controlling a computer}\label{fig:compco}
\end{figure*}

\begin{figure*}[h]
\centering
\includegraphics[width=\textwidth]{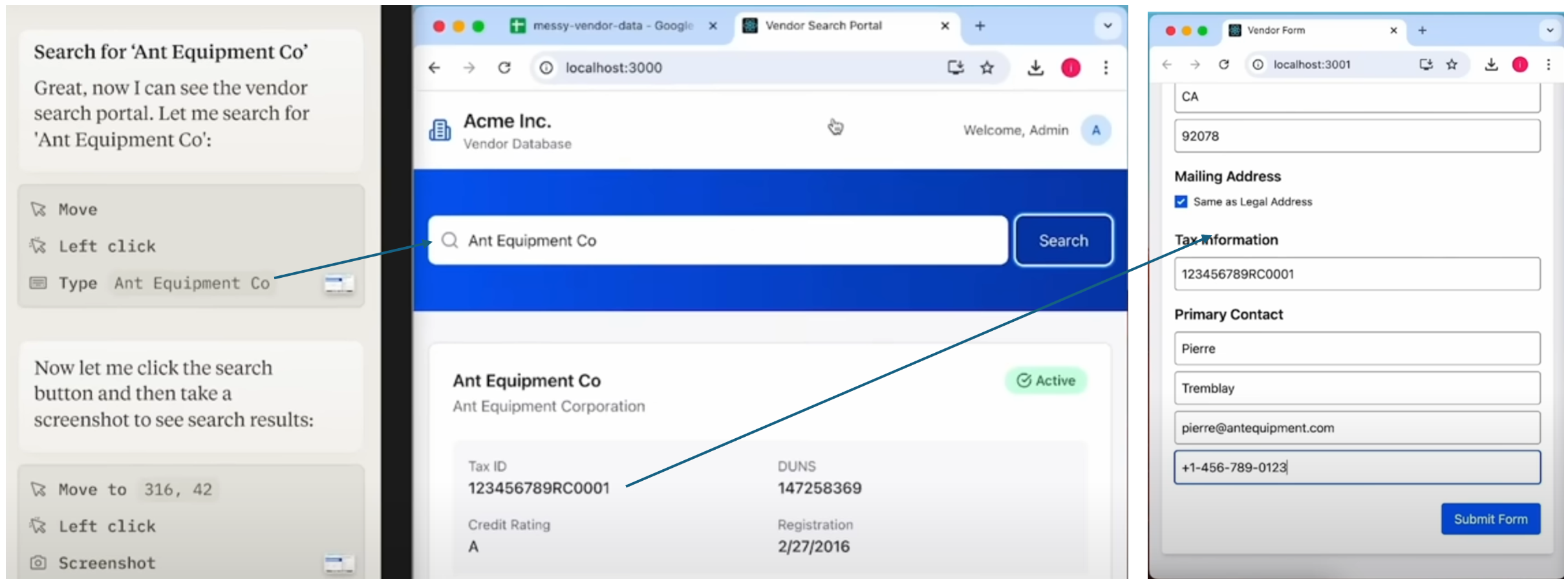}
\caption{(Anthropic's) agent controlling a computer through mouse, keyboard and sensing through screenshots as illustrated in the left panel}\label{fig:compco2}
\end{figure*}

\subsubsection{Interacting in (Virtual) Worlds} Often agents operate in restricted or digital environments though attempts have been made to consider general environments moving towards models that consider the world in its entirety or at least narrow tasks involving the physical world (or a model of it). For example, agents have also been employed in specialized simulation environments for robotics~\citep{wan24la}.

Multi-modal models that are trained on text, images and videos might constitute a potential world model~\citep{ge24wo}. That is, such models should, for example, given an image of the current state (such as a plane) and a textual description of an action (the plane will land soon) predict the next state (showing a plane that has landed).~\cite{xia23} deploys an agent in a model of a virtual home, which simulates the physical world. The agent performs random exploration and goal-oriented planning to gain experience, which are used to fine-tune an LLM. The fine-tuned LLM performs significantly better than much larger non-fine-tuned LLMs. Due to their extensive knowledge LLMs can be used as both a world model and a reasoning agent engaging in planning~\citep{hao23rea}. Here the reasoning agent builds a reasoning tree suggesting actions to solve a task under the guidance of the world model providing rewards. 

While ordinary generative vision-language models allow to textually describe images and generate images from text. Agentic AI seeks to go beyond that by inferring actions given images and text. In particular,~\cite{bro23} developed a vision-language-action model by training a vision-language model on trajectories of robots containing both their observations and actions visual language tasks such as visual question answering.

\smallskip
\noindent\textbf{Motion planing and 3D: }
While such approaches might improve significantly on prior work for the considered scenario, they still fall short in many aspects and tasks that are innate for humans. For example, they cannot perform 3D spatial reasoning such as estimating distances or size differences between objects without extensive training on this task~\citep{che24vlm}.

The idea to leverage LLMs for motion planning, i.e., planning driving trajectories has also been explored~\citep{mao23}. They described a visual scene in text by specifying objects and their coordinates, which allowed a language model to process and reason upon the geometrical problem and output the motion trajectory.

\smallskip
\noindent\textbf{Agents in computer games: } Agents are commonly evaluated in computer games as they can be seen as open-worlds. 
~\cite{not23emb} uses two phases to play the strategy game "MineCraft": i) an LLM to plan the behavior of an RL agent by defining subgoals, ii) an RL agent learning a policy for each subgoal and updating the LLM's world model ~\citep{wan23dep} proposed an interactive planning approach for MineCraft based on describe, explain, plan, and select to yield feasible plans and reduce errors through self-explanation.

\smallskip
\noindent\textbf{Interacting with browsers and computers: }
The idea to control a computer or at least the browser has gained increased attention. Commercial companies like OpenAI and Anthropic~\citep{Ant24} have already released agents performing simple tasks as illustrated in Figure~\ref{fig:compco}, where the user prompt is addressed by controlling the mouse, keyboard and taking screenshots~\ref{fig:compco2}. Furthermore, a number of benchmarks where agents are supposed to solve tasks. To this end, multi-modal models can be used to extract information relevant to control the system from screenshots, e.g., as done with Omniparser V2~\citep{yu25om}. 

\begin{figure*}[h]
\centering
\includegraphics[width=\textwidth]{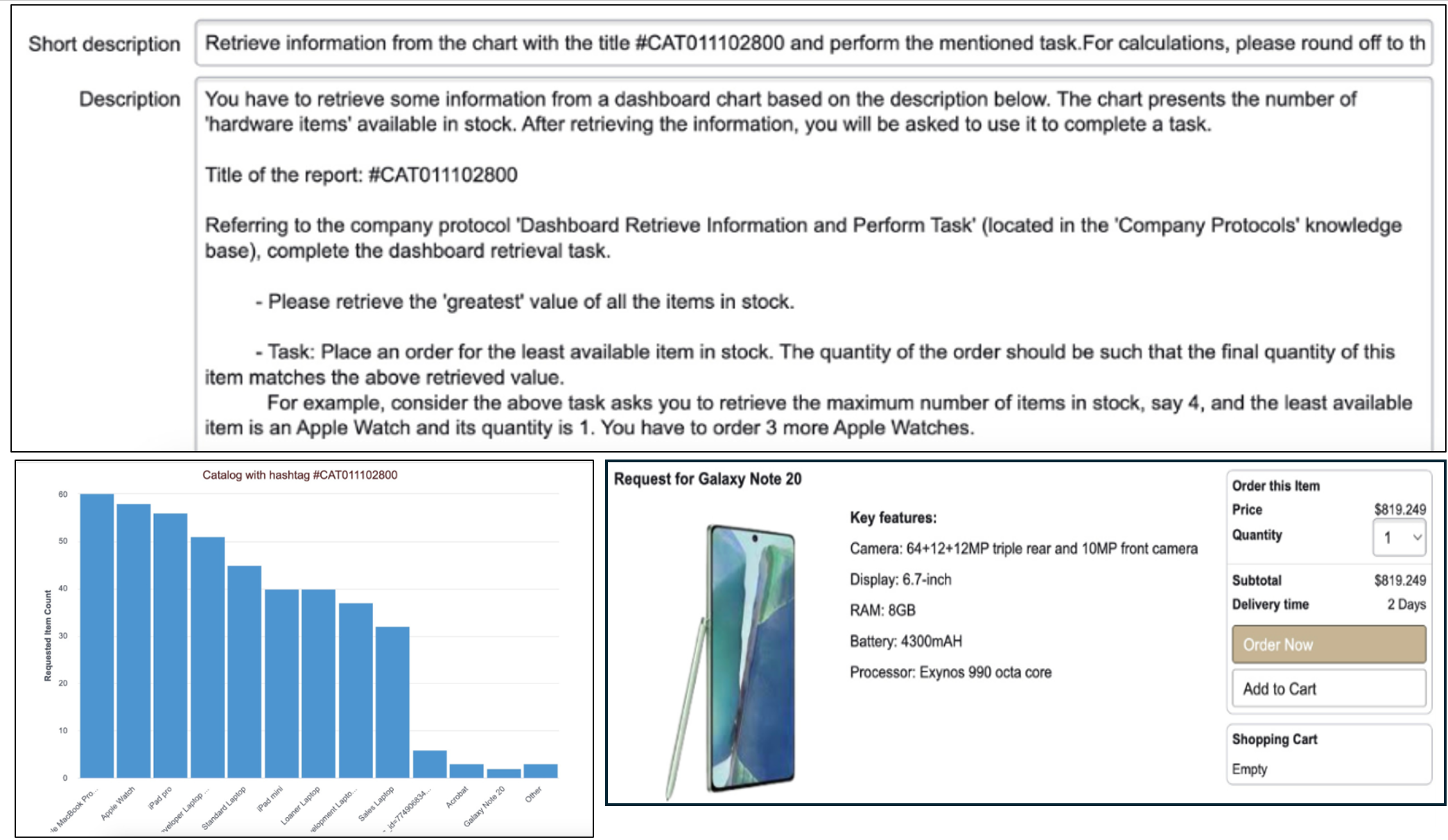}
\caption{Task from Workarena++~\citep{boi24}. Given a ticket asking to refill inventory below four items (top panel), the system must understand a chart to identify items to order(left lower panel) and interact with an ordering page (right lower panel).}\label{fig:arena}
\end{figure*}

A web-browser can be seen as a tool that provides access to web-services but also as an environment to solve tasks. Systems with reasoning and planning capabilities can be enhanced with acting capabilities, e.g.,~\cite{zho23la}, to make use of web-browsers~\citep{drou24,zhou23web,boi24}. However, as of now performing ``relatively'' simple tasks for humans, e.g., related to online shopping remains a challenge for AI agents~\citep{jin24sho}. More concretely, on the Workarena++ benchmark~\citep{boi24} humans score close to 100\% while LLMs such GPT-4o and LLama3 score close to 0\%. An example task is shown in Figure \ref{fig:arena}.

While the idea of agents controlling a computer with all its software and any further tuning towards agent usage is appealing, designing specialized agent friendly interfaces can improve an agent's success rate as demonstrated for software engineering~\citep{yan24swe}.

\smallskip
\noindent\textbf{Interacting with tools:} Conceptually, tool usage (and access to external memory like databases) could also be viewed as environmental interaction. However, we treat both as integral components of the Agentic system. First, memory and tool usage often do not require a long sequence repeating perceiving, reasoning, acting, but rather tool usage is a single API call, e.g., to a database. Tool usage is often learnt in a supervised manner, e.g., by using samples of problem descriptions and the corresponding API calls. Second, we view an agent as being equipped with at least basic capabilities to use a tool. Reasoning might be needed to apply them successfully for a task if no supervised fine-tuning is performed. But if an agent lacks knowledge on tool usage, an agent can often rely on documentation, usage samples, and obtain immediate feedback, which tends to limit the needed interaction. However, especially for complex tools where no training data exists, a reinforcement learning setting might be the better conceptual match.

\subsubsection{Interacting with Humans}
Humans can assist, collaborate, or oversee the agent.

\noindent\textbf{Scientific Processes: } The AI scientist~\citep{lu24ai} and follow-up works~\citep{gri25} implement an end-to-end LLM driven scientific discovery process including idea generation, experiment iteration and paper write-up showing that it can generate 100s of medium-quality papers within a week. 
Deep Research by OpenAI and Google also support researchers by generating research reports, e.g., searching and summarizing works from a particular area or about a particular topic using multiple rounds of search and analysis. In particular, these commercial tools also often ask humans for clarifications as the research process takes a significant amount of time due to a large computational demand. In the academic world, human-agent collaboration for research has been discussed in~\citep{ifa25aut}.
More specialized research agents have been proposed, e.g., for drug discovery~\citep{liu24dru}. Specialized agents can perform highly domain specific tasks, such as simulating molecules relevant to chemistry and modeling ecosystems as needed in biology~\citep{che24exp}.

\smallskip
\noindent\textbf{Agents as Assistants: }
Agents can be used to support the execution of tasks but also for improving skills, e.g., practicing negotiations~\citep{sch23neg}. Despite an extensive number of already deployed applications, the optimal design of agents engaging with humans is still subject to research. For instance, recently back-channeling as an active listening strategy, where the LLM would utter phrases like "really?" or "Wow" at appropriate times led to higher conversational engagement~\citep{jan24}.

Many areas aside from the aforementioned ``scientific discovery'' such as law, finance, psychology, education, medicine and military also benefit from Agentic AI with numerous applications~\citep{che24exp}. For example, an Agentic AI workflow has been proposed to translate formal medical reports into patient-accessible reports reducing errors and hallucination through reflection~\citep{sud24ag}. Counseling agents for students being bullied have also been assessed~\citep{pau24pee} mainly by comparing existing commercial LLMs. For more medical examples consult~\citep{wan25me}. Agentic AI has also been suggested as ethical counsel in the practice of law~\citep{o24ag}. In finance, multi-agent systems have been proposed for decision making~\citep{yu24fin} as well as for trading~\citep{xia24tra}.

\subsubsection{Interacting with Other Agents}
Agents can assume various roles: autonomously simulating aspects of society or systems, acting as integrated analytical components like critics or evaluators within reasoning processes, or collaborating independently to solve complex tasks through coordinated efforts. 

\smallskip
\noindent\textbf{Agents for Simulation:} Agents have been used for societal and economic simulations, such as modeling macroeconomic dynamics with diverse interacting agents~\citep{li24eco}. Agents have also been employed to simulate disease spread, such as during the COVID-19 pandemic~\citep{wil23ep}. In recommender systems, agents simulate both users and items to model interactions and improve recommendation quality~\citep{zha24cf}. Agents have also represented countries in simulations of historical wars~\citep{hua23war}. 

\smallskip
\noindent\textbf{Agents as Analytical Components:} Having two LLM agents debate under the moderation of an LLM judge has fostered more divergent thinking~\citep{lia23en} and improved summarization showin in benchmarks~\citep{cha23}. However, agents have also shown to reduce diversity by converging towards human-like polarization~\citep{piao25em}. Additionally, incorrect or manipulated knowledge can spread rapidly within agent networks~\citep{ju24flo}. 

\smallskip
\noindent\textbf{Agents as Independent Collaborators:} To solve user-defined tasks,~\cite{li23ca} proposed using role-playing agents, such as a Python programmer and a stock trader collaborating to develop a stock trading bot. Although these agents work together autonomously, the paper suggests introducing critics to provide feedback and improve outcomes. The paper also emphasizes the importance of carefully crafted initial prompts and precise task specifications, potentially generated by an LLM.~\cite{hao25cha} proposes using a hierarchical structure of agents led by a central leader. In each layer, messages from previous layers are aggregated, producing an averaging effect that stabilizes responses. The leader provides feedback that subordinate agents use for improvement.

\subsection{Specifying and Evaluating Agentic AI}
A critical aspect of any system is its specification and ensuring that the system conforms to the specification and broader goals such as legal compliance and efficiency. 
\subsubsection{Specifying an Agent}
Defining an agent typically involves describing identity information (e.g., name, age, personality), motivational drivers, professional roles~\citep{par23gen}, tool permissions, delegation rights, workflows, and interaction behavior with other agents—see Figure \ref{fig:crewSample}. Although agent descriptions can be elaborate, simple descriptions often suffice—for instance, in a multi-agent programming setting, agents were defined using only a professional role, a goal, and constraints~\citep{hon23}. The definition also depends on the intended level of autonomy \ref{tab:aut}; agents with partial autonomy—whether due to technical limitations or legal reasons—may require consent checkpoints where humans review outcomes and planned future actions~\citep{sin22con}.

\smallskip
\noindent\textbf{Agent Design: Manual and Automatic (Data-Driven or Model-Based):} Agents can be designed either manually or automatically. Automatic design can involve fitting to data or through model optimization involving, e.g., an evaluation function.~\cite{hu24au} evaluates the automated design of agentic systems. This approach requires specifying a search space of potential agents, a search algorithm to explore the space, and an evaluation function to assess the quality of agents. A meta-agent programs new agents by generating code, potentially based on archived prior agents, and employs a reflection process to ensure novelty. Evolutionary algorithms using cross-over and mutation have also been used for automatic agent creation~\citep{yua24evo}. Data fitting can be achieved using few-shot prompting with examples or through sampling techniques. For example, to emulate human participants in a user study, agent profiles can be generated by sampling traits from a distribution~\citep{arg23out}.~\cite{wan23rec} manually defined a few agent profiles, which were subsequently expanded into a larger agent pool with the assistance of an LLM.

\begin{figure*}[h]
\centering
\includegraphics[width=0.9\textwidth]{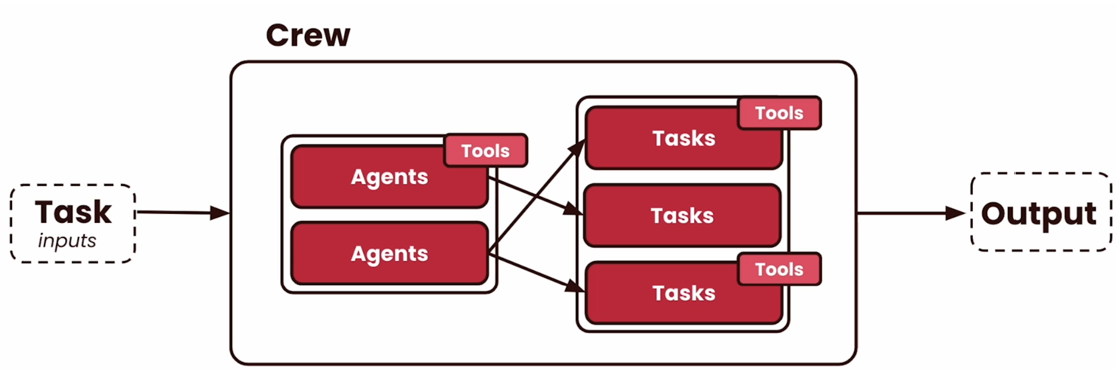}
\caption{Specifying a multi agents system using agents, tasks and their interconnection as abstractions specified by a user (Figure from \citep{mou24cre})}\label{fig:agfram}
\end{figure*}

\begin{figure*}[h]
\centering
\includegraphics[width=\textwidth]{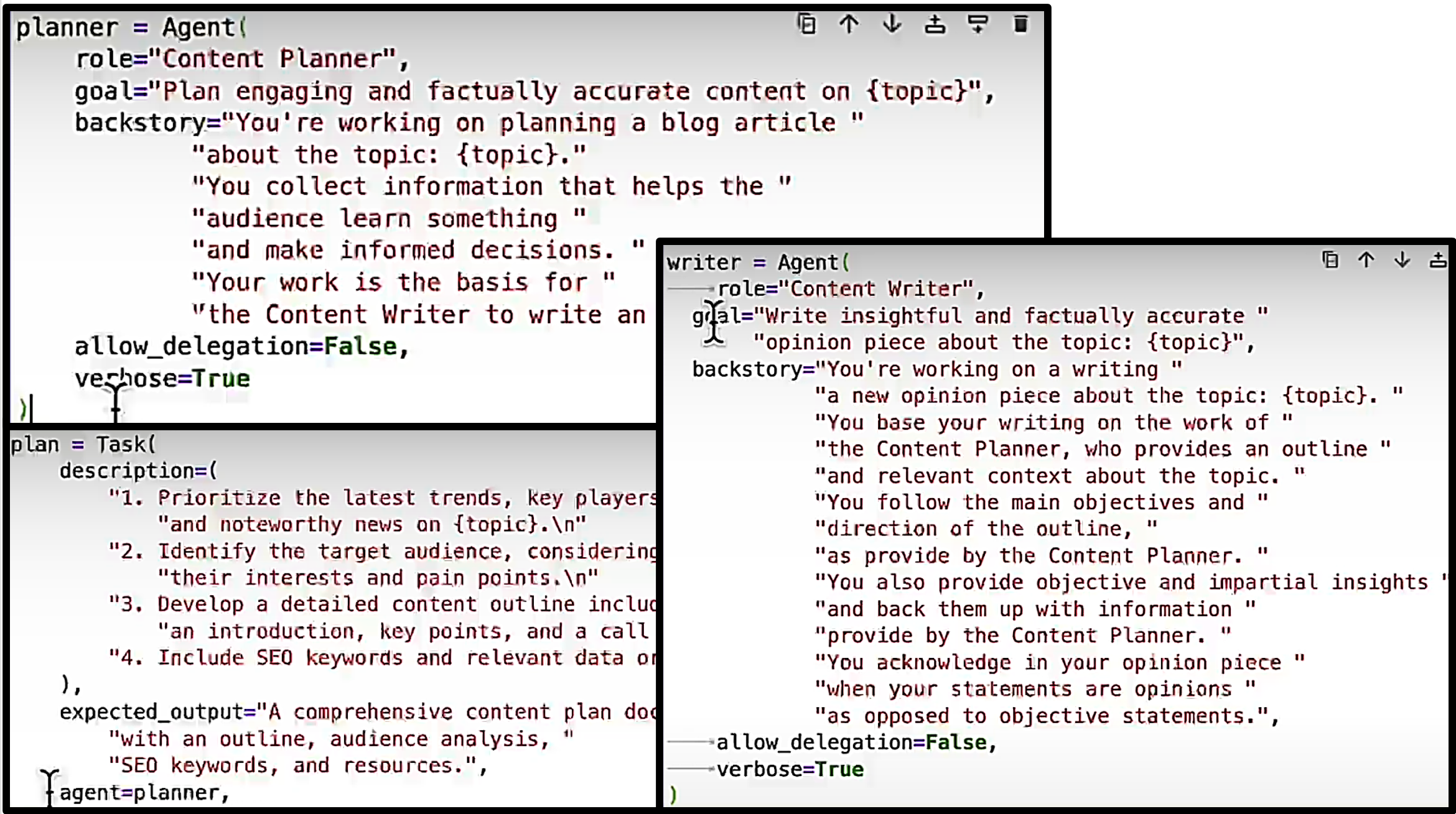}
\caption{Example definition of a content planner agent, the content planning task and the writer agent providing procedural guidance including interaction with other agents  (Figure from~\citep{mou24cre})}\label{fig:crewSample}
\end{figure*}

\subsubsection{Specifying Multi-Agents}
Agentic systems often comprise multiple agents. Using agents with different personas—potentially based on the same underlying LLM—can achieve better outcomes than chain-of-thought (CoT) prompting for tasks like creative writing and trivia~\citep{wan23un}, motivating the adoption of multiple agents. Multi-agent systems can be defined in various ways but generally follow similar structural patterns. Agents and tasks are defined independently, allowing flexible assignment of agents to tasks, as illustrated in Figure \ref{fig:agfram}. Both agents and tasks can be flexibly specified through textual descriptions, similar to conventional prompting. An agent description mirrors that of a single-agent system, specifying goals, a role (or persona) characterizing attributes (e.g., empathetic behavior), allowed tool usage and delegation, constraints, and procedural task guidance. A task description may include expected outputs and procedural guidance, including instructions for agent interactions, as shown in Figure \ref{fig:agfram}.

Multi-agent systems form distributed architectures where communication patterns and (self-)organization strategies, such as symmetry breaking~\citep{bar16}, are crucial.

\smallskip
\noindent\textbf{Key dimensions of Multi-Agents System: }
We provide essential dimensions for multi-agents systems synthesized from works that provide a more detailed discussion (e.g.,~\cite{che24exp,guo24la,li24sur})

\begin{itemize}[leftmargin=1.5em]

  \item \textbf{Homogeneous vs. Heterogeneous:}  Homogeneous agents can be exact replicas, meaning the underlying foundation model as well as the agent description are identical. Heterogeneous models can be based on different models, subject to different agent instructions. 
  
  \item \textbf{Cooperative vs. Non-cooperative:} Cooperative agents pursue a shared goal and may use argumentative debate styles to enhance results~\citep{du23}.

  \item \textbf{Communication structure:}
 Communication can be classified as \textit{push}—where messages are sent unsolicited—or \textit{pull}—where agents request information explicitly. Communication can also be \textit{centralized}—through a single agent or shared message pool~\citep{hon23}—or \textit{decentralized}, where agents interact freely without a central node.  Standardization of communication like the agent to agent protocol\cite{goo25a2a} facilitate interaction across diverse agents.

  \item \textbf{Organization (Flat vs. Hierarchical):} Agents can be organized in hierarchical settings, e.g.,~\cite{hao25cha}, or in a flat manner.  For example, ~\cite{zha24cha} employs multiple workers handling parts of a task (sequentially) and a manager aggregating results.

  \item \textbf{Learning:} Agents may learn independently in decentralized systems or share experiences centrally to accelerate collective improvement.

  \item \textbf{Communication content:} Communication is primarily text-based; cooperative systems often package instructions, goals, state descriptions, action histories, and dialogue histories into messages~\citep{zha23bui}. Agentic systems often mimic real-world development by featuring domain-expert agents that collaborate through discussions, potentially adopting varied communication styles~\citep{yam25}. 
\end{itemize}

\subsubsection{Evaluating Agentic AI}
Evaluating Agentic AI is challenging. Agents can behave non-deterministic with complex interactions and require a much longer time-span to complete tasks than typical for GenAI.
Agents can exhibit a wide range of capabilities and desiderata that require evaluation. We outline key decision domains but refer readers to specialized surveys for further details~\citep{yeh25}. Capabilities span general LLM skills like long-text understanding and instruction following, as well as agent-specific skills such as planning, learning from interactions, error handling, tool usage, and spatial reasoning~\citep{wu23sma}. Desiderata include task performance~\citep{wu23sma}, failure awareness (recognizing impending failure), efficient tool usage, harmfulness~\citep{and24har}, and computational efficiency—for instance,~\cite{arc2} enforced efficiency by setting computational limits for tasks. Security is also a key concern as agents come with their own threat model \citep{nar25sec}. Beyond foundational capabilities, agents can also be evaluated on domain-specific competencies. For instance, evaluating agents simulating humans in user studies has involved manual comparisons with real users and sanity checks that vary names and gender to account for prompt sensitivity and model shortcomings~\citep{ahe23us}.

\smallskip
\noindent\textbf{Metrics} can include generic measures such as completion rate (fraction of tasks completed) or task success rate~\citep{wan23dep}, possibly supplemented by solution quality scores, self-aware failure rates (fraction of tasks where agents signal failure before failing), and tool call accuracy. In games, evaluation scores could reflect deviations from optimal actions (e.g., in rock-paper-scissors) or achievements awarded by the game engine~\citep{wu23sma}.

\smallskip
\noindent\textbf{Benchmarks }  designed for Generative AI, such as the well-known MMLU benchmark~\citep{hen21mml}, can also be applied to Agentic AI; MMLU has been extended with more complex questions to better evaluate reasoning abilities~\citep{wan24mml}. Various datasets exist for evaluating (or training) Agentic AI, often focusing on specific areas such as tool usage (over 14,000 REST APIs~\citep{qin23}), robot learning~\citep{wal23br}, assistant-user task pairs~\citep{li23ca}, programming~\citep{hen21mea}, social game playing~\citep{aka23pl}, and simulating human behavior in user studies~\citep{ahe23us}. Frameworks are also available to facilitate task evaluation.~\cite{lin23ag} offers infrastructure enabling researchers to easily design custom evaluation tasks. Games are frequently used for evaluation, ranging from simple games like rock-paper-scissors to complex environments like Minecraft~\citep{wu23sma}. Several benchmarks target evaluating agents in web environments~\citep{drou24,pan24web}, focusing on completing basic human tasks~\citep{drou24}.


\section{Towards AGI?}
\subsection{Challenges Moving Towards AGI}

Many challenges found in GenAI also apply to Agentic AI, though some become either more pronounced or mitigated. They hinder the further evolution to Artificial General Intelligence (AGI). These challenges also posit research opportunities.

\smallskip
\noindent\textbf{Errors:} Although reasoning reduces errors compared to Generative AI, as shown in benchmarks (Figure \ref{fig:mmlu}), the complexity and length of tasks in Agentic AI increase the risk of cumulative errors across steps~\citep{sha23ag}.

\smallskip
\noindent\textbf{Robustness:} Foundation models are known to be highly prompt sensitive~\citep{schn24f}. In turn, agents relying on such models generating their own prompts as part of the reasoning process, are also highly sensitive to original prompts and potentially also to minor changes in the environment.

\smallskip
\noindent\textbf{Interpretability:} Agentic AI emphasizes step-by-step reasoning presented in readable text, enhancing interpretability. However, Chain-of-Thought (CoT) reasoning may not always be faithful to the underlying decision process~\citep{tur23}. Moreover, GenAI~\citep{sch24expl} and deep learning broadly~\citep{lon24exp} continue to struggle with explainability. Additionally, Agentic AI is more complex than GenAI because it integrates components like planning algorithms, external memory, and tool usage. 

\smallskip
\noindent\textbf{Dynamic and Complex Environments:} Agents are designed to operate in non-deterministic, complex environments. They interact with changing environments, select tools, and dynamically collaborate with other agents, sometimes in parallel.

\smallskip
\noindent\textbf{Observability:} The environment might not be fully observable, e.g., we cannot gather all information about our world. This is a common assumption in RL~\citep{rus21}. Furthermore, agents’ behaviors and tool operations may not be fully transparent. For example, agents might access tools via APIs whose internal mechanisms are opaque.

\smallskip
\noindent\textbf{Safety and Security:} Agents are vulnerable to harmful behaviors, either through deliberate attacks~\citep{and24har} or systemic shortcomings. This risk is higher for Agentic AI systems compared to GenAI, due to their increased interaction in less controlled environments. For instance, agent capabilities like tool invocation can be exploited. Furthermore, multi-modal agents also face attacks on each modality, e.g., foundation models can be jailbroken through adversarial images~\citep{qi24vis}. Privacy concerns escalate as agents share information with other agents and tools. Even seemingly minor information—such as an agent's role as a psychological counselor—can represent a privacy breach. 

\smallskip
\noindent\textbf{Unpredictable Emergent Behaviors:}
LLMs have shown remarkable emergent behavior due to scale, e.g., in-context learning\cite{bro20} or chain-of-thought reasoning \cite{wei22cha}. Such behaviors are difficult to foresee in many ways -- in fact, many capabilities of models might even be dormant and waiting to be discovered. While for LLMs most can be deemed positive -- or at least posing no immediate threat, this might change with ``agent swarms'' interacting with the real-world, e.g., such agent swarms could put democracies at risk at an unprecedented scale as elaborated in a paper by a large number of reputable researchers \cite{schr25}. 

\smallskip
\noindent\textbf{Evaluation:} Evaluating Agentic AI—assessing reliability, task performance, and potential harms—remains challenging and incomplete~\citep{sha23ag}. 

\smallskip
\noindent\textbf{Human Alignment:} Agents may engage in unforeseen or unethical actions in pursuit of their goals, contrary to human intent. 

\smallskip
\noindent\textbf{Controlling and Monitoring Agents:} Controlling agents is challenging. Even when human or independent approvals are required for agent actions, anticipating the consequences of approvals or disapprovals remains difficult~\citep{sha23ag}. Because agents are dynamic, they may adopt unpredictable behaviors to achieve their goals. 

\smallskip
\noindent\textbf{Resource Allocation and Management:} Managing resources, such as computational access, is more demanding as agents dynamically consume varying and potentially arbitrary amounts of information~\citep{che24exp}.

\subsection{Agentic AI to AGI?}
AGI refers to AI that transcends narrow domains and can generalize to new, unfamiliar situations. Current AGI benchmarks, such as~\citep{arc2}, emphasize novel tasks requiring learning from few examples and limited computation, rendering brute-force approaches impractical. Agentic AI enables novel applications beyond those possible with GenAI. Tasks requiring more complex reasoning and interactions become more and more feasible. Currently, agents are restricted to relatively simple tasks due to technological and regulatory constraints, but this is expected to evolve. Agentic AI may represent the next step toward Artificial General Intelligence (AGI), potentially equaling or surpassing human intelligence. Incremental technological advances and scaling of existing Agentic AI systems might be sufficient to achieve AGI. However, progress may stall due to barriers like limited availability of training data~\citep{mok25}. Some researchers argue that mere scaling is insufficient, and fundamentally new approaches are required~\citep{ben23sca}. Consequently, the timeline for achieving AGI remains uncertain, as evidenced by the wide range of predictions from respected researchers. For instance, in 2018, Ray Kurzweil predicted a 50\% chance of human-level AI by 2029, while Stuart Russell estimated 50–70 years, Yann LeCun 50–100 years, and Rodney Brooks approximately 180 years~\citep{ford18arc}.

\noindent\textbf{What happens if AGI is reached? } Even decades before the current AI boom, there were speculations about the consequences of achieving AGI~\citep{kur05,bos14su}. It is arguably difficult to predict the outcomes once AI surpasses human intelligence. Researchers like Stuart Russell suggest that once AI reaches a "Kindergarten level," its improvement could accelerate at more than an exponential rate~\citep{ford18arc}. In particular, AGI has been said to posit an existential risk to humanity~\citep{bos14su} posing challenges with respect to control and value alignment. According to the instrumental convergence thesis, AI systems might independently pursue goals like self-preservation and resource acquisition, potentially clashing with human interests. These existential risks go far beyond more tangible economic risks such as unemployment and potential erosion of wages~\citep{bos14su}.

\section{Methodology and Related Work}

\subsection{Methodology and Scope} 
\noindent\textbf{Scope and Target Audience: }
Our focus is on contrasting Agentic AI and GenAI and highlighting key innovations in the transition, targeting a broad audience of academics and industry professionals. The basic characteristics distinguishing GenAI from Agentic AI are intended to be accessible to a wide audience, while recent innovations in Agentic AI are tailored for a more technically focused readership. To this end, we adopt a high-level capability perspective aimed at general interest. We assume readers have basic familiarity with deep learning and Generative AI concepts, including foundation models and prompt engineering, as covered in prior works (e.g.,~\cite{schn24f}). In-depth technical knowledge, such as of the transformer architecture, is not our focus, as it is covered extensively in other works (e.g.,~\cite{sch24com}). While we address all relevant areas of Agentic AI, we refer readers to other works for more extensive coverage of specific aspects.

\noindent\textbf{Research Methodology: } We primarily followed the literature review methodology outlined by~\citep{woh14gui}, incorporating several innovations to address the rapid growth of research in this field. Specifically, we began by conducting a meta-survey, searching Google Scholar for surveys on “Agentic AI,” “LLM agents,” and “Generative AI.” This approach served three goals: (i) identifying key papers for forward and backward search (i.e., snowballing~\citep{woh14gui}), (ii) building on existing works to ensure conceptual completeness, and (iii) improving prior works by synthesizing different viewpoints and identifying gaps. We assessed more than 30 surveys, but identified five surveys on Agentic AI~\citep{wan24su,xi25ris,ach25,pla25,che24exp} and two on Generative AI~\citep{zha23com,man24cha}, which were chosen due to quality, recency, comprehensive and non-overlap with other chosen surveys. The Agentic AI surveys were significantly more influential in shaping our work. We conducted an initial read-through of these manuscripts. We then developed an outline for our paper, including chapter structures and basic content, unifying prior works while omitting less relevant aspects and introducing novel ones. In a second iteration, we revisited the surveys to identify gaps in our conceptualization, refining our framework primarily through backward search. In a third iteration, we conducted forward searches, focusing on citations and publications in top venues like NIPS, ICLR, and ICML, while also considering new arxiv.org papers. We restricted our search to works from 2024 onward (up to October 2026), as older works were assumed to be captured in the selected surveys. Additionally, we performed targeted searches on Google Scholar for each subchapter, emphasizing recent surveys on topics like evaluation, reasoning, planning, and RAG. These sources were used to further refine the structure and content of our work. To ensure the inclusion of the most recent results, we incorporated works from arxiv.org and selected blog posts from reputable sources such as OpenAI and Huggingface. However, these additional sources were included only after basic quality assessments. Also, note that the initial version from 2025 was updated in two subsequent iterations.

\subsection{Related Surveys} 

As stated in the research methodology, we relate most strongly five surveys on ``Agentic AI'' and ``LLM Agents''. Key differentiators include: (i) None of the surveys focused on distinguishing GenAI from Agentic AI, which forms the first part of our survey. (ii) None offered a clear and multi-angled definition of Agentic AI. (iii) None discussed autonomy levels and the motivation for Agentic AI—including a contrasting perspective to AGI—in comparable detail. The second part of our survey overlaps more with existing works, but we expand on important practical aspects such as (iv) defining agents and memory characteristics, and (v) offer a different conceptual framework. We believe that offering multiple, diverse perspectives on Agentic AI is highly valuable. We now discuss these surveys in greater depth and highlight how they differ from our work:

~\cite{wan24su}, surveying “LLM-based autonomous agents,” adopts an architecture-centric view describing key components but lacks a detailed discussion on defining agents. Our work is more capability-focused and differs significantly in its conceptualization. Additionally, we emphasize distinguishing GenAI from Agentic AI. 

~\cite{xi25ris} adopts a conceptualization emphasizing brain, perception, and action, whereas we focus on reasoning and interaction. They explore the historical notion of agents and the motivation for using LLMs, while we emphasize the differentiation between GenAI and Agentic AI.

~\cite{ach25} surveys “Agentic AI” by comparing it to traditional AI, including early rule-based systems. In our view, the traditional AI era began in the 1960s and ended with the breakthroughs of Generative AI, notably GPT-2/3 around 2019. While their comparison is highly valuable, it is of less interest towards readers already familiar with GenAI such as ChatGPT. Our survey targets these readers by starting with Generative AI and not covering the entire history of AI. As a result, our conceptualization differs significantly—for example, Table 1 in~\citep{ach25} and our Table \ref{tab:compCap} share only one common aspect. 

~\cite{pla25} surveys “Agentic LLMs.” Our work differs by contrasting Agentic AI directly against GenAI. Our conceptualization also differs: at the highest level, we emphasize two rather than three core capabilities; additionally, we view retrieval augmentation not as reasoning (as in Figure 3 of~\citep{pla25}), but rather as interaction with a tool, aligning with earlier works like~\citep{lew20}. Basic reasoning forms like chain-of-thought~\citep{wei22cha} do not require retrieval, and even more advanced reasoning typically only necessitates large context windows. We offer a broader set of motivations for Agentic AI (compare Table \ref{tab:limgen} in our work to Chapter 1.4 in~\citep{pla25}) and discuss agent definitions in greater depth. 

~\cite{che24exp} surveys “LLM-based agents” by comparing RL and LLM agents, whereas we contrast Agentic AI against GenAI. Our conceptualization differs at multiple levels: at a high level, we focus on novel AI agent capabilities like reasoning and interaction, while~\citep{che24exp} adheres closely to traditional RL agent concepts.

~\cite{sap25ai} compares ``AI agents and Agentic AI'' also touching on differences to GenAI. It is more focused on describing applications (like \cite{kha26cri}), while we provide a more fine granular resolution of conceptual differences between generative and agentic AI discuss more extensively reasoning, interaction, and almost entirely missing aspects such as self-improving agents \cite{opcl26sel}. We account more for the historic and gradual evolution from generative to agentic AI (see, e.g., Figure \ref{fig:genToAg}). We also have different conceptual views, e.g., in our view aligned with modern AI agents (e.g., \cite{opcl26sel}) an AI agent is not limited to a single task execution (see Table 8 \cite{sap25ai}).

\section{Conclusion}
Agentic AI is a major paradigm shift beyond Generative AI by introducing reasoning, interaction, and autonomy at a new scale. Our survey and conceptualization systematically contrasts Agentic AI and GenAI from multiple perspectives. We cover its technical foundations, practical specification, and open challenges. As Agentic AI evolves, understanding its capabilities, risks, and the nuances of agent specification becomes essential for both advancing research and ensuring responsible deployment.

\section*{Declarations}

\subsection*{Conflict of Interest}
The authors declare that they have no conflict of interest.

\subsection*{Ethical Approval}
Not applicable.

\subsection*{Informed Consent}
Not applicable.

\subsection*{Funding}
No funding was received for conducting this study.

\subsection*{Data Availability Statement}
Data sharing is not applicable to this article as no datasets were generated or analyzed during the current study. All references are included in the paper.

\subsection*{Author Contributions}
Not applicable. There is just one, which did everything.

\subsection*{Acknowledgments}
Not applicable.

\bibliography{refs}
\end{document}